\documentclass[journal,twoside]{{IEEEtran}}
\usepackage{amsfonts}
\usepackage{amssymb}
\usepackage{graphicx}
\usepackage{subcaption}
\usepackage{color}
\usepackage{multirow}
\usepackage[ruled,vlined]{algorithm2e}

\begin{document}
\title{Spatio-Temporal Graph Representation Learning for Fraudster Group Detection 
}
%
%
\author{
Saeedreza Shehnepoor*,
Roberto Togneri,
Wei Liu,
Mohammed Bennamoun,
\thanks{S. Shehnepoor (*corresponding author), 
R. Togneri,
M. Bennamoun, and
W. Liu is with the University of Western Australia, Perth, Australia.
emails: \{saeedreza.shehnepoor@research.uwa.edu.au, roberto.togneri@uwa.edu.au, wei.liu@uwa.edu.au, mohammed.bennamoun@uwa.edu.au.\}}
}
%
%
%
\maketitle              
\begin{abstract}
Motivated by potential financial gain, companies may hire fraudster groups to write fake reviews to either demote competitors or promote their own businesses. Such groups are considerably more successful in misleading customers, as people are more likely to be influenced by the opinion of a large group. To detect such groups, a common model is to represent fraudster groups static networks, consequently overlooking the longitudinal behavior of a reviewer thus the dynamics of co-review relations among reviewers in a group. Hence, these approaches are incapable of 
excluding outlier reviewers, 
which are fraudsters intentionally camouflaging themselves in a group and genuine reviewers happen to co-review in fraudster groups. To address this issue, in this work, we propose to first capitalise on the effectiveness of the HIN-RNN in both reviewers' representation learning while capturing the collaboration between reviewers, we first utilize the HIN-RNN to model the co-review relations of reviewers in a group in a fixed time window of 28 days. We refer to this as spatial relation learning representation to signify the generalisability of this work to other networked scenarios. Then we use an RNN on the spatial relations to predict the spatio-temporal relations of reviewers in the group. 
In the third step, a Graph Convolution Network (GCN) refines the reviewers' vector representations using these predicted relations. These refined representations are then used to remove outlier reviewers. The average of the remaining reviewers' representation is then fed to a simple fully connected layer to predict if the group is a fraudster group or not. Exhaustive experiments of the proposed approach showed a 5\% (4\%), 12\% (5\%), 12\% (5\%) improvement over three of the most recent approaches on precision, recall, and F1-value over the Yelp (Amazon) dataset, respectively. 
\end{abstract}

\section{Introduction}
\label{sec:intro}   
\begin{figure}[ht]
\centering
\includegraphics[width=\linewidth]{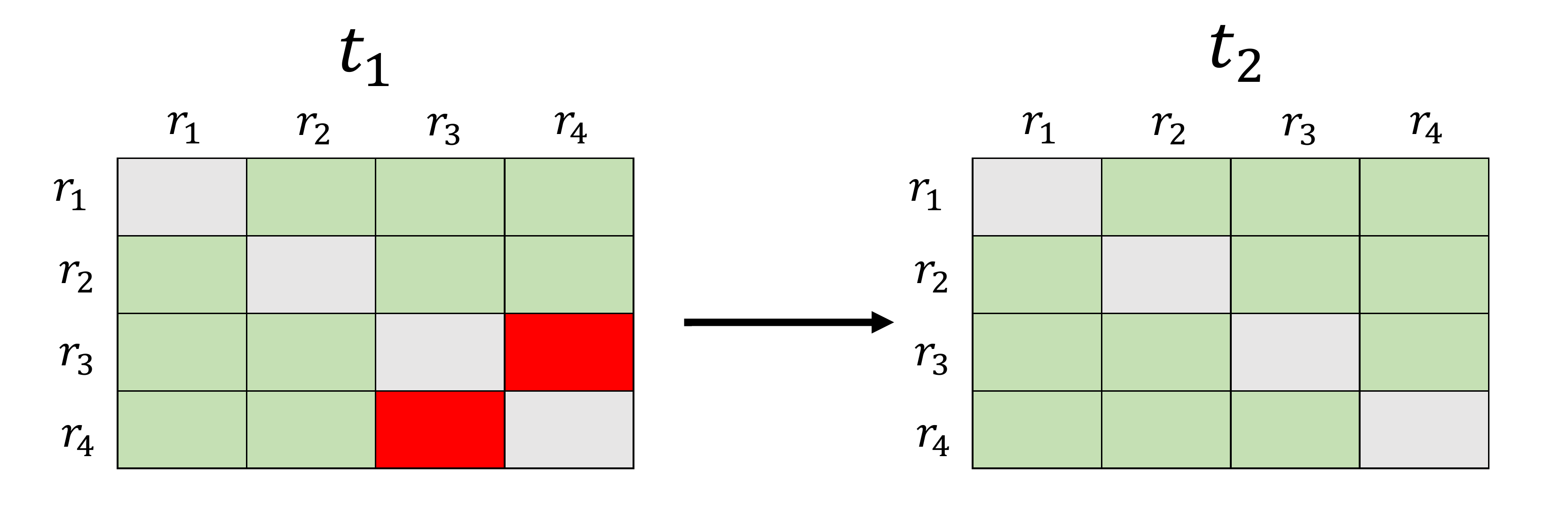}
\caption{A toy example of a co-review matrix demonstrating the behavior change of fraudsters in a fraudster group (green means genuine reviews and red means fraud ones). In time step $t_1$, reviewers $r_4,r_3$ co-review an item with fraud reviews. In the next time step ($t_2$) the reviewers are camouflaged.
} 
\label{fig:tempo-toy}
\end{figure}

Modern consumers’ significant uptake of online shopping, has created a flourishing e-commerce environment, including the increasing reliance on product review recommendations such as those found on Yelp and Amazon. This has provided a fertile ground for fraud reviewers to deliberately mislead consumers through manipulated reviews. To increase their impression on users, fraudsters may work in teams, thus forming fraudster groups, to collectively attack (or promote) certain products or services. Group fraud may serve different purposes such as manipulating the semantics of reviews, distributing the overall workload and avoiding detection through temporal behavior manipulation ~\cite{Mukherjee2012,Lingyun2018,Ye2015}. The fraud related research area is currently dominated mainly by individual fraudster detection~\cite{Jindal2008,MukherjeeV0G13,dfraud} and fraud review detection~\cite{shehne2017,Shebuit2015,shehnepoor2020scoregan}. Recent years have seen an increased research effort in detecting fraudster groups~\cite{Allah2013,JI2020454,Zhang2020,Xu2019}. It is widely accepted that individual fraudsters can cause significant damages to businesses, fraudster groups may be even more damaging because of their coordinated and considerate volume of fraud reviews that they can collectively produce. 
Fraudster groups also are much more difficult to detect compared to individual fraudsters. Each group fraudster can camouflage more easily by controlling his/her relation with other group members such that no single fraudster stands out~\cite{Mukherjee2012,Ye2015}. 
In other words, a group fraudster can escape detection by avoiding certain relations with other group reviewers or develop multiple relations with genuine reviewers. As a member of a community~\cite{chen2010game}, group fraudsters can also manipulate their relations over time to appear to be genuine reviewers and, as such the group appears authentic and avoids detection, as illustrated in Fig.~\ref{fig:tempo-toy}.\\ 
The algorithms for fraudster group detection are dominated by two categories: Frequent Itemset Mining (FIM) based~\cite{Allah2013,Xu2019} approaches or graph-based~\cite{JI2020454,Zhang2020} approaches. An FIM-based algorithm generally follows a two-step process: \textbf{first,} candidate groups are determined based on the same set of items (itemset) reviewed by the reviewers. \textbf{Then} the candidate groups are ranked based on the probability of being a fraudster or genuine group. Note that the candidate group refers to a group of users with a possible fraudulent collaboration. Graph-based algorithms, on the other hand, employ graph partition or clustering algorithms~\cite{Ye2015} to provide a representation of the reviewers to determine the candidate groups' probability of being either fraudster or genuine. However, the approaches in both categories suffer from significant drawbacks. 

\textbf{First,} the most recent approaches (Ji \textit{et al.}~\cite{JI2020454}, Zhang \textit{et al.}~\cite{Zhang2020}, and Shehnepoor \textit{et al.}~\cite{shehnepoor2021hinrnn}) overlooked the temporal nature of fraudster groups, as illustrated in Fig.~\ref{fig:tempo-toy}. Fraudster groups can take advantages of such a limitation to manipulate relations and mislead detection algorithms.  
\begin{figure*}
\centering
\includegraphics[width=\linewidth]{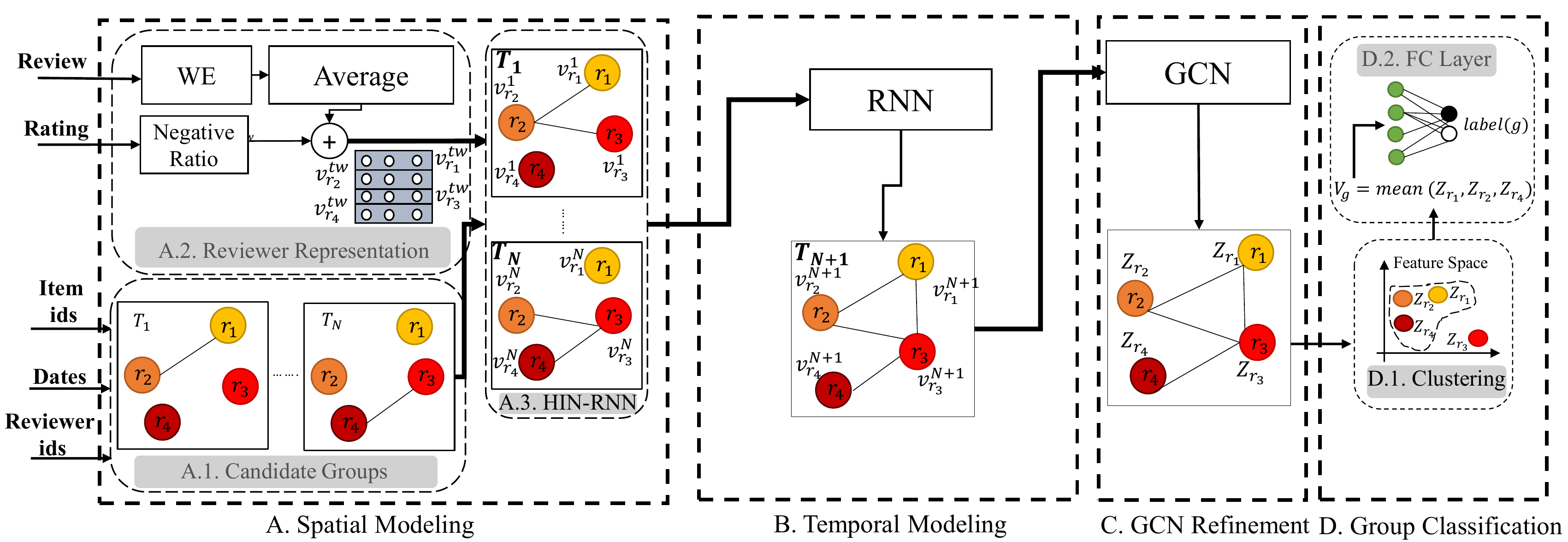}
\caption{The framework of our proposed approach. WE is the Word Embedding. Note that the links in each time window are representing the connection between \textbf{reviewers of a group} in that time window, and no link means there is no connection in that time window, while they form a connection, either temporally in other time windows or spatially.
} 
\label{fig:framework}
\end{figure*}

\textbf{Second,} previous approaches are limited to a confined task of using either members' relations or reviewers' vector representations in the final classification. A better approach should consider all the available information to jointly learn both the  semantic representation and the behavioral representation.

\textbf{Third,} current approaches on fraudster group detection are incapable of modeling ``outlier" members of a group. In social review platforms, apart from purposely forming a fraudster group, reviewers may also form genuine groups due to similar interests. Note that the genuine groups are not necessarily formed and in this study the genuine groups refer to genuine reviewers writing similar reviews for a same set of items. However, members in a fraudster group could make genuine reviews, and vice versa, members in a genuine group may write fake reviews. Such outliers; i.e., fraudster(s) in a genuine group or genuine reviewer(s) in a fraudster group; pollute the group, creating another level of challenges for fraudster group detection.  
The most recent approach of fraudster group detection by Shehenepoor \textit{et al.}~\cite{shehnepoor2021hinrnn} excluded such outliers by removing the users with the least connection with other reviewers. However, such an approach overlooks the importance of the joint representation in covering different aspects of a reviewer's activities in a social platform. Joint representations have shown to be effective in different fraud detection tasks~\cite{Mukherjee2013,shehne2017,Shebuit2015}, and hence are hypothetically helpful in removing the outliers, which has been verified effective by the results of this paper (See Sec. \ref{sec:reviewer-removal-effect}).
Removing outliers with the least connection results in some improvement, however, such a removal approach is still not effective enough in removing outlier reviewers, resulting in a higher False Positive (FP) rate and False Negative (FN) rate. 

Fig.~\ref{fig:framework} uses a toy example of a group with four reviewers to illustrate our proposed algorithm. Our proposed spatial-temporal architecture is a four module pipeline which continuously refines the initial reviewer representation for the downstream fraudster group detection task. The first module is responsible for the ``spatial" relation between reviewers. This is not the spatial distance in the geographical sense, but a derived social closeness based on whether two reviewers have co-reviewed. We define co-review as a relation between two reviewers when they reviewed the same item with the same rating and similar semantics in a specified window of times. For example from Fig.~\ref{fig:framework}. A, during $T_1$, reviewer $r_1$ and $r_2$ have a co-review relationship whereas during $T_N$, $r_3$ and $r_4$ have co-reviewed.  We take advantage of the HIN-RNN~\cite{shehnepoor2021hinrnn} to model the co-review relation between reviewers, as it has been proven to be effective in utilizing spatial relations between reviewers. 
HIN-RNN~\cite{shehnepoor2021hinrnn} is an RNN used to model the relation between nodes in a graph considering the nodes' heterogeneity. The heterogeneity is obtained through a fine-tuning of the semantics extracted using word embedding techniques based on the reviewer's type (fraudster/genuine). 
Given the reviews, reviewer ids, item ids, ratings, and dates as the input, the proposed spatial modeling comprises three steps: 
\begin{enumerate}
    \item First, the candidate groups in each time window are determined. 
    \item With groups determined, reviews are used to extract the Sum of Word Embeddings (SoWEs) for the reviewers in each time window as the semantic representation of the reviewer.
    \item The HIN-RNN refines the candidate groups from the first step using the semantic representation extracted in the second step. The time windows ($t_i$) are then utilized in the reviewer representation step to extract the SoWE for the reviewer $i\in\{1,2,3,4\}$ in a corresponding time window ($V^{TW}_{r_i}$). The relations between reviewers in groups for different time windows are then refined with the reviewers' representation by the HIN-RNN~\cite{shehnepoor2021hinrnn} to output the spatial relations. The HIN-RNN realizes a possible spatial correlation between $r_2$ and $r_3$ in the $1^{st}$ and $N^{th}$ time window. 
\end{enumerate}
The output of the spatial modeling step are subgraphs with refined relations between reviewers in a group in each time window.

The second module employs a simple Recurrent Neural Network (RNN) to model the relations between the reviewers throughout different time windows, given the spatial relations from the previous spatial module. 
The RNN predicts a possible temporal relation between, say, $r_1$ and $r_3$ (from the possible similar temporal activities depicted in Fig.~\ref{fig:framework}. B). This step addresses the previous approaches' limitation in temporal relation modelling, by encoding the reviewers' temporal relations. The output of this step is the collaboration matrix for each group at the $N+1^{th}$ time window.

In the third module, a GCN (Graph Convolutional Network) is used to ensure that the reviewers' effects on each other are captured. Hence, the GCN will refine the representation of the reviewers based on the captured spatio-temporal relations and the labels of each individual reviewer. For example, from Fig.~\ref{fig:framework}. C, reviewers $r_1,r_2,r_4$ have the type, while $r_3$ has a different one. So, the GCN provides further refinement based on the reviewers' representation, the spatio-temporal relation, and the labels of the reviewers. The output of this step are the groups with the reviewers' representation refined based on the spatio-temporal relations that were captured in the previous steps. 

After the refinement in the third step, we apply the K-means clustering algorithm in the fourth step to recognize the outlier reviewers in each group based on the representations. As a result of such refinement, the proposed approach can exclude outlier reviewers in this step, 
although reviewer $r_3$ developed multiple relations with other reviewers to escape the detection (as shown by Fig.~\ref{fig:framework}. D1). This ensures the inclusion of a joint representation in removing outliers (one of the key limitations of previous studies). The average of the remaining reviewers' representation is then fed to a Fully-Connected (FC) layer for the final classification of the group (Fig.~\ref{fig:framework}. D2).
\\

We can summarize our contributions as follows:
\begin{itemize}
    \item For the first time, we propose an approach to model the spatio-temporal relations between reviewers in the groups. Such an approach not only produces a more accurate modeling of the reviewers' relations through spatial relations, but also considers the possibility of fraudsters' camouflage through the cover-up of reviews with a group of genuine reviewers or the unintentional genuine reviewers' involvement in a fraudster group activity over time (See. Sec.~\ref{sec:temporal-modeling-effectiveness}). The proposed approach outperforms the state-of-the-art approaches (Ji \textit{et al.}~\cite{JI2020454}, Zhang \textit{et al.}~\cite{Zhang2020}, Shehnepoor \textit{et al.}~\cite{shehnepoor2021hinrnn})  by 5\% (4\%), 12\% (5\%), 12 (5\%)\% based on the precision, recall, and F1-value on the Yelp (Amazon) dataset, respectively (See Sec.~\ref{sec:main-results}).    
    \item
    Given the spatio-temporal relations between the reviewers in the groups, we use a Graph Convolutional Network (GCN) to refine the initial semantic-behavioral representation (SoWE + NR) of reviewers. This step discriminates further between genuine and fraudster reviewers in terms of their representations. The results show that the GCN effectively improves the performance of the proposed approach by 11\% and 5\% in terms of recall and F1-value on the Yelp dataset, respectively (See Sec.~\ref{sec:refinement-effectiveness}). 
    \item With the refined representations, we propose a new approach to remove outlier reviewers from the group using the well-known K-means clustering algorithm. The results show that the proposed clustering algorithm outperforms the threshold based removal approach proposed by Ji \textit{et al.}~\cite{JI2020454} and the minimum connection removal approach proposed by Shehnepoor \textit{et al.}~\cite{shehnepoor2021hinrnn} (See Sec.~\ref{sec:reviewer-removal-effect}).   
\end{itemize}
The rest of the paper is structured as follows. In Section \ref{sec:related-works}, we discuss the related work. In Section \ref{sec:proposed-approach}, we introduce our methodology. In Section \ref{sec:experimental-evaluation}, we provide our experimental evaluation. We conclude the paper with an outlook to future work in Section \ref{sec:conclusion}.

\section{Related Works}
\label{sec:related-works}
Previous studies on fraudster group detection are generally categorized based on the strategy to determine the initial group formations and includes two subcategories: Frequent Itemset Mining (FIM) and Graph-based algorithms. 
\subsubsection{Frequent Itemset Mining} 
\label{sec:FIM}
Approaches in this category determine the initial groups based on an assumption that the reviewers with the same set of reviewed items (itemset) form a possible collaboration, thus a possible group~\cite{Agrawal94}. 
Allahbakhsh \textit{et al.}~\cite{Allah2013} extended the FIM concept and defined a new detector called the biclique detector. The groups' formation was initialized based on the biclique concept: reviewers writing reviews with the same rating on the same group of items. After the group formation initialization, handcrafted features such as Group Rating Value Similarity (GVS) and Group Rating Time Similarity (GTS) were extracted. A scoring function was applied to the extracted features to determine the probability of a group being fraudster or genuine. The proposed approach yielded a precision of 75\% for the fraudster group detection on the Yelp dataset.
Xu \textit{et al.}~\cite{Xu2019} also relied on the FIM concept to determine the candidate groups, where reviewers with at least two reviews on at least three co-reviewed products formed a collaboration. A measure called homogeneity-based-Collusive Behavior Measure (h-CBM) was proposed using the targeted item, rating, temporal traits, and the reviewer activity. To score the groups, an unsupervised scoring model called Latent Collusive Model (LCM) was employed. The approach proposed by Xu \textit{et al.} showed a precision of 85\% on the Yelp dataset.
To overcome the limitations of previous approaches, Shehnepoor \textit{et al.}~\cite{shehnepoor2021hinrnn} proposed HIN-RNN, incorporating the semantic representation of the reviewers to refine the relations. For that purpose, Shehnepoor \textit{et al.} first determined the candidate groups based on the FIM concept and then extracted the representation of the reviewers using a Continuous Bag of Words (CBoW). Next, the HIN-RNN was proposed to refine the initial representation. Finally, to reduce the effects of outlier reviewers, the reviewers with minimum relations were removed. The representations of the remaining reviewers were used to predict the probability of the group being a fraudster or not. HIN-RNN achieved a precision of 81\% on the Yelp dataset.\\
\subsubsection{Graph-based} 
\begin{figure}[ht]
\centering
\includegraphics[width=\linewidth]{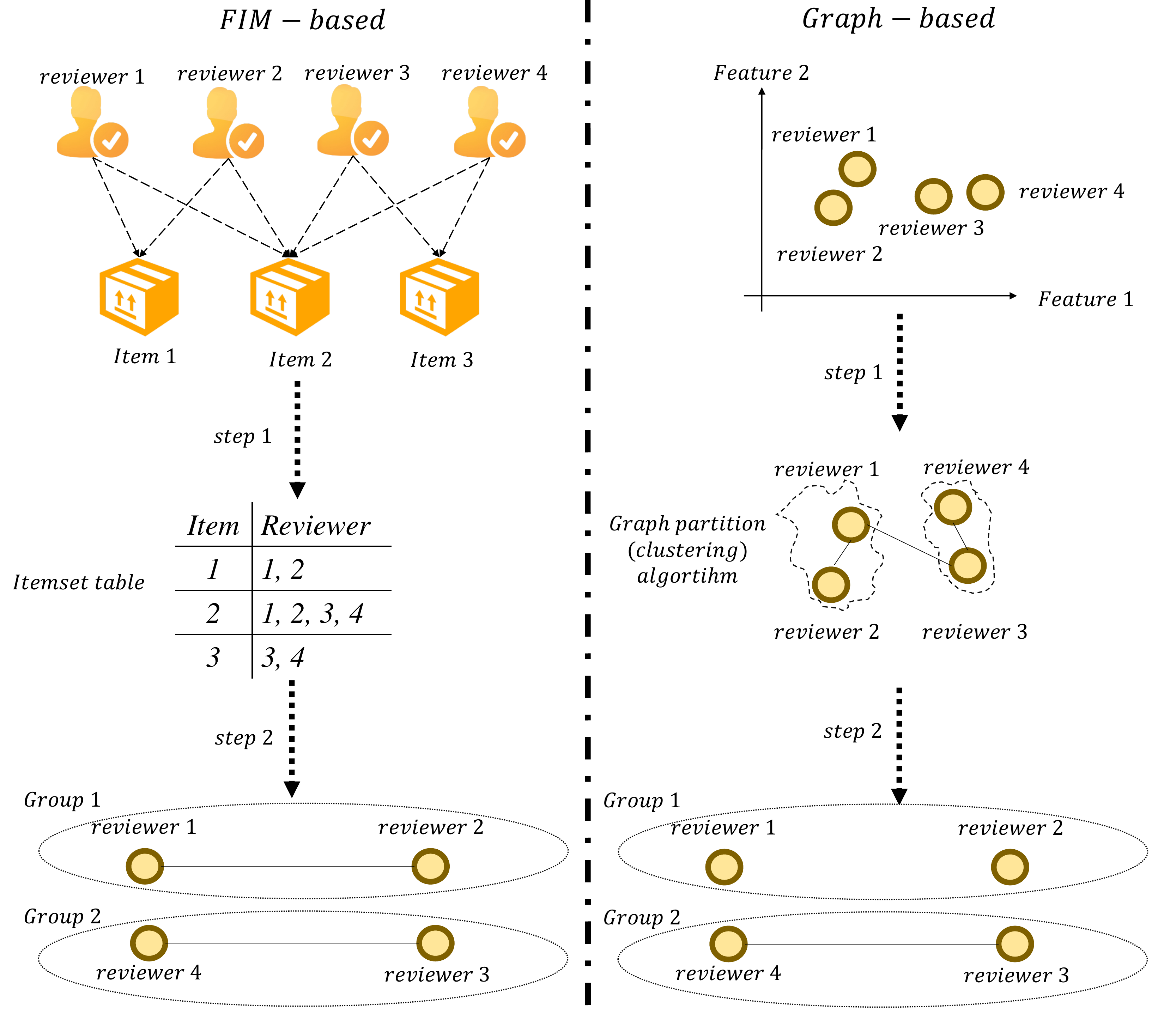}
\caption{FIM-based Vs. graph-based approaches~\cite{shehnepoor2021hinrnn}.} 
\label{fig:fim-vs-graph}
\end{figure}

Graph-based approaches use graph partition algorithms or clustering methods to determine the group formation based on the similarity between reviewers' representation. 
Ji \textit{et al.}~\cite{JI2020454} proposed an approach with a focus on products as the main target of the fraudster groups. Ji \textit{et al.} claimed that considering the products as the focal point will overcome the limitation of FIM based approaches in concentrating only on reviewers.
To have a representation for each group, seven handcrafted features (Group Rating Deviation, Group Size, Group Review Tightness, Group One Day Reviews, Group Extreme Rating Ratio, Group Co-Activeness, and Group Co-Active Review Ratio) were computed. Next, to demonstrate the effectiveness of the item representation in fraud detection, three product related features (Product Rating Distribution, Product Average Rating Distribution, and Suspicious Score) were extracted. Finally, six individual fraudster features (Ratio of Extreme Rating, Rating Deviation, The most Reviews One-day, Review Time Interval, Account Duration, and Active Time Interval Reviews) were used to obtain a single representation for each reviewer in a group.
Targeted items were then scored based on item related features. A Kernel Density Estimation (KDE) was used to compute the burstiness for items. Outlier reviewers were also removed based on a threshold applied to each reviewer representation. The proposed approach by Ji \textit{et al.} showed a precision of 83\% on the Yelp dataset. However, both temporal behavioral clues and reviewer representation refinement were overlooked in modeling the relations between reviewers. Zhang \textit{et al.}~\cite{Zhang2020} proposed a framework with three steps. In the first step, the similarity between reviewers' ratings and their reviewed items was used to build a graph with reviewers as nodes and the similarities as the edges in the next step.
Finally, to obtain the final prediction, a label propagation algorithm was proposed based on the propagation intensity and an automatic filtering mechanism. The proposed approach by Zhang \textit{et al.} provided a precision of 70\% on the Yelp dataset. To demonstrate a comparison of FIM-based and graph-based approaches, an overall view of both approaches is shown in Fig.~\ref{fig:fim-vs-graph}.
\subsubsection{Research Gap} 
Although FIM based approaches and the graph-based ones provide insight for fraudster group detection, they overlook the temporal clues of the reviewers' behavior in a fraudster group. They also do not consider the reviewers' relations in a group to determine the reviewer's representation. Hence, such approaches also suffer from a limitation to consider the possibility of fraudster (genuine) reviewers' involvement in genuine (fraudster) groups. In this research, we therefore devise a new approach to \textbf{1)} model temporal relations (co-reviewing) between reviewers to predict the collaboration matrix, \textbf{2)} refine the relations between the reviewers in a group based on their spatio-temporal relations, and \textbf{3)} use the relations to refine the representations of the reviewers. In previous works the outlier reviewers were also removed based on a static threshold or based on the number of relations, resulting in inaccurate pruning, easily manipulated by camouflage. We recognize this problem and utilize a clustering algorithm to remove the outlier reviewers. 
\section{Proposed Approach}
\label{sec:proposed-approach}
\textit{\textbf{Problem Definition:}} Let's consider a time stamped review as a tuple $\langle r, t, i, d\rangle$ where reviewer $r\in R$ wrote text $t\in T$ on item $i\in I$, on date $d\in D$. The goal of fraudster group detection is to obtain subgraphs $R^g\subset R$ and classify them into either fraud or genuine. 
\subsection{Spatial Modeling}
\label{sec:spatio-modeling}
In this step, we aim to model the spatial relations between reviewers in the groups throughout different time windows with no overlaps. 
\subsubsection{Candidate Groups}
\label{sec:group-construction}
To increase their impact, fraudster group members are likely to write coordinated fraud reviews in a shorter time frame as compared with genuine reviewers~\cite{Mukherjee2012,Mukherjee2013}. Similar to previous studies, the reviewers with the same set of co-reviewed items with the same ratings in a period of 28 days establish an initial relation in a group~\cite{JI2020454,Zhang2020,Wang2018}, thus a candidate group.
The members of the initial candidate group grows to include reviewers who co-reviewed in future time windows of 28 days. This resulted in $N$ networks of all reviewers who have ever co-reviewed in any of the $N$ time windows of 28 days. Two reviewers will be connected only when they co-reviewed in the specific time window $TW_i$, where the duration of time window is 28, which can be set to other duration to meet domain specific needs.  
So the output of this step is subgraphs of possible groups captured in $N$ time windows (as shown in Fig.~\ref{fig:framework}.A). We refer to such subgraphs as \textit{co-review} networks, as  a representation of a candidate group in a specific time window. 


\subsubsection{Reviewer Representation}
\label{sec:reviewer-rep}
\begin{figure}
\centering
\includegraphics[width=\linewidth]{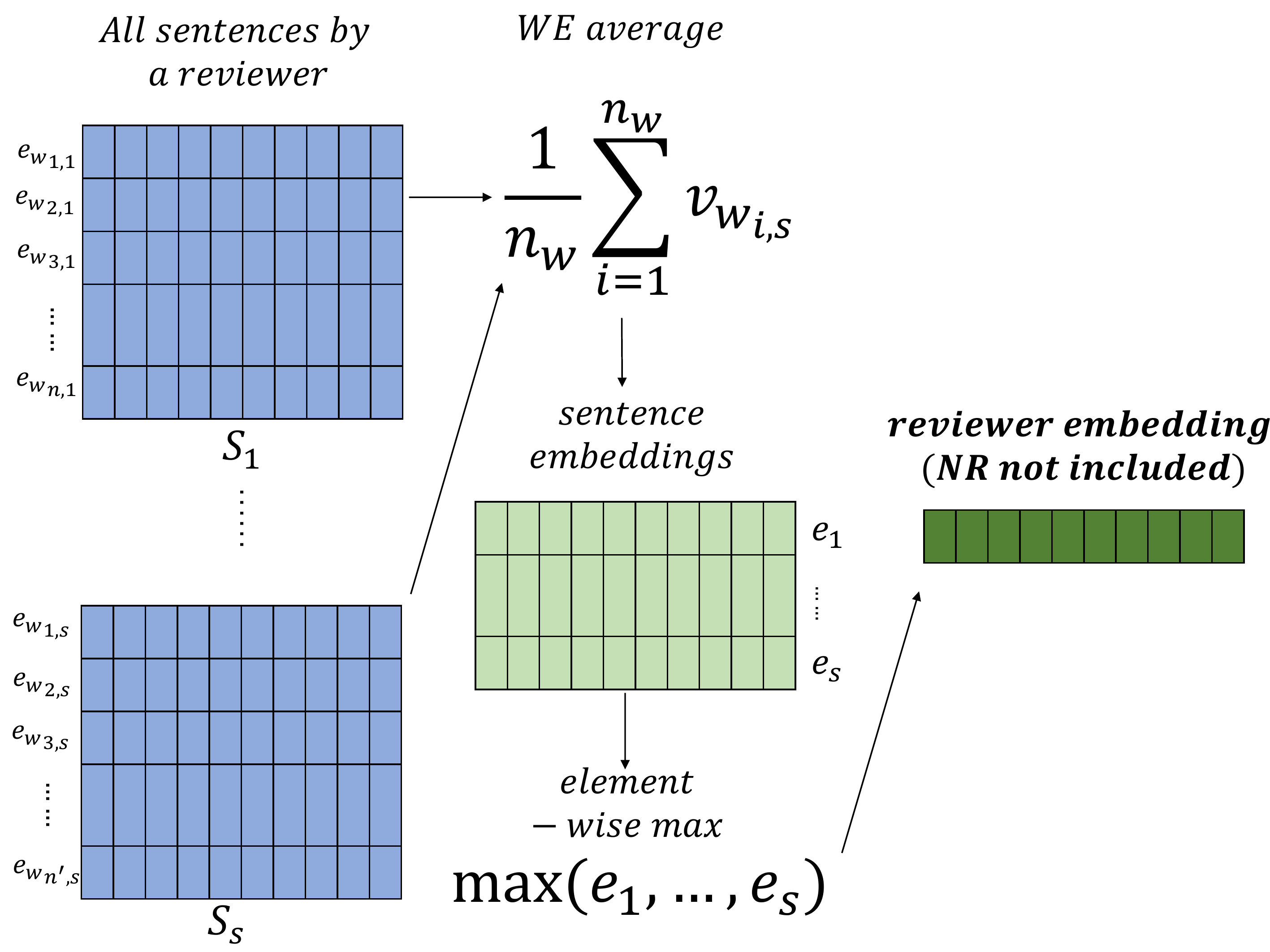}
\caption{The structure of the semantic representation steps for a reviewer using SoWE. 
} 
\label{fig:reviwer-rep}
\end{figure}

In this step, reviews for each reviewer are aggregated and then split into sentences. 
With the promising performance of Word Embedding techniques in fraud detection~\cite{Ren:2017:NND:3043977.3044107}, each word in the sentences is initialized with a pre-trained word embedding, $e_{w_i}\in \mathcal{R}^F$, as the embedding for $i^{th}$ word ($w_i$) from a vector space with dimensionality of $F$. 
The Sum of Word Embeddings (SoWE) of a reviewer's review text is used as a semantic representation of each reviewer. SoWE refers to the simple linear function of aggregating the embeddings of words to represent a sentence, a reviewer, or a group, shown to be effective in different domains, such as Ren \textit{et al.}~\cite{Ren:2017:NND:3043977.3044107}, White \textit{et al.}~\cite{White2018}, and Lyndon \textit{et al.}~\cite{Lyndon2015}. Therefore, to obtain a SoWE for a sentence we applied an element-wise average to WEs of a sentence:
\begin{equation}
\label{eq:sen-rep}
e_s =\frac{1}{n_w}\sum_{i=1}^{n_w}v_{w_{i,s}}
\end{equation}
where the $n_w, v_{w_{i,s}}$ is the number of the words in the sentence, and the corresponding word embedding of word $i$ in the sentence $s$, respectively. Finally, we apply a max (as max-pooling) operator to obtain a vector representation for the reviewer.
The process is depicted in Fig.~\ref{fig:reviwer-rep}.

Next, we extract a behavioral indicator for each reviewer, namely the Negative Ratio (NR)~\cite{Shebuit2015}:
\begin{equation}
\label{eq:nr}
NR = \frac{\sum_{rating=1}^{2}n_{rating}}{N}
\end{equation}
where the $n_{rating}$ is the number of reviews of a reviewer with specific ratings ($rating$)  in the range  of  1-5 (5 is the highest rating), and $N$ is the total number of reviews by the reviewer. The concatenation forms the final vector representation of reviewer $r$ as $v_r$.

\subsubsection{Spatial Modeling Through HIN-RNN}
\label{sec:HIN-RNN}
\begin{figure}
\centering
\includegraphics[width=\linewidth]{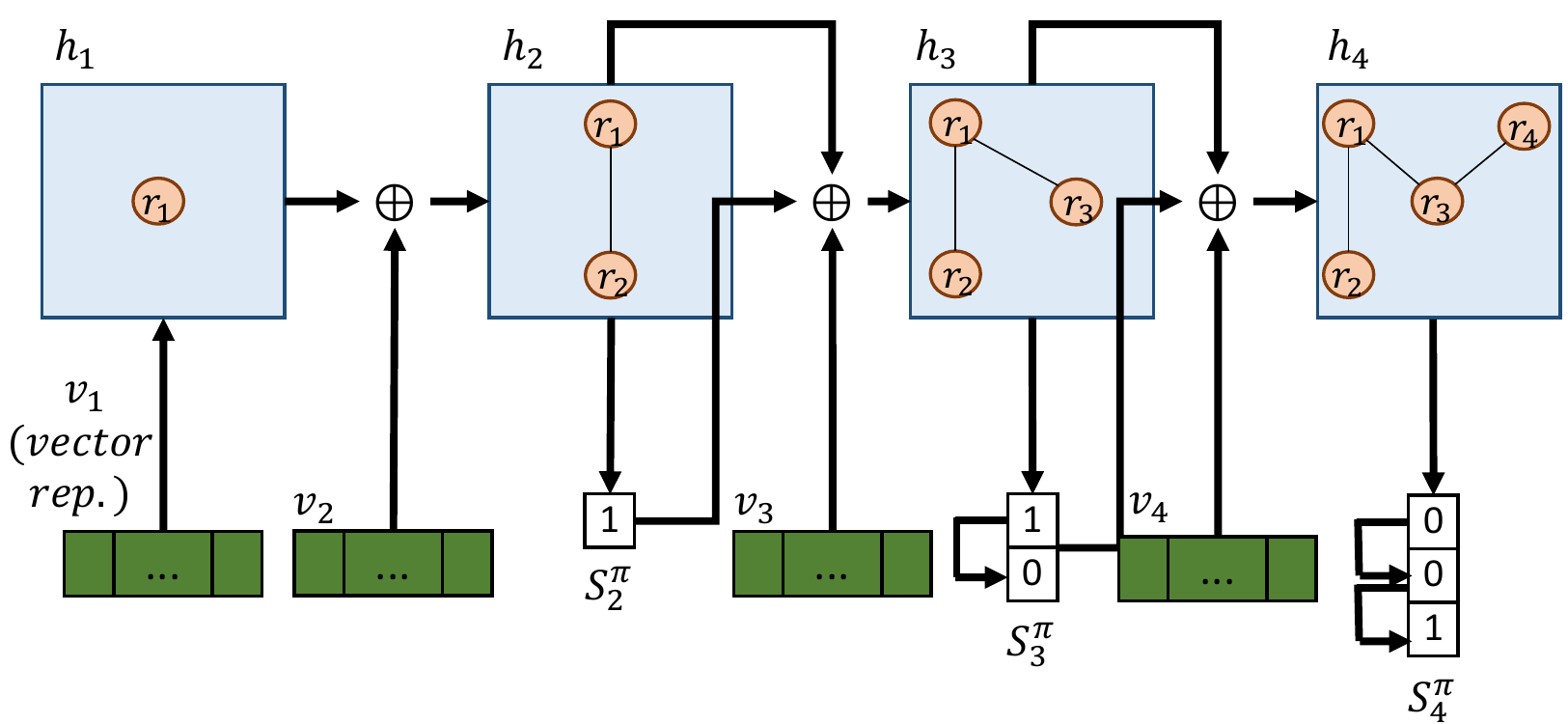}
\caption{The structure of the HIN-RNN~\cite{shehnepoor2021hinrnn}. The HIN-RNN uses the representation of each reviewer, the graph state, and the collaboration vector of the previous node to generate the collaboration vector for the current reviewer. Note that $\oplus$ is the concatenation operator. $h_i$ represents state of the group in $i^{th}$ step, while $v_i$ is the representation of $i^{th}$ reviewer.} 
\label{fig:HIN-RNN}
\end{figure}

Given the effectiveness of the HIN-RNN~\cite{shehnepoor2021hinrnn} in taking into account the reviewers' long-range dependencies, we employ the HIN-RNN to model the relations between reviewers in different time windows. HIN-RNN takes the co-review networks as subgraphs, alongside the reviewers' representation, and outputs the refined relations between the reviewers based on the reviewers' representations. Therefore, first the subgraphs are mapped to corresponding adjacency matrices called ``collaboration matrix" using a node ordering $\pi$ ($\pi(r_1), \dots, \pi(r_i), \dots, \pi(r_n)$), with $r$ representing the reviewer and $n$ is the number of reviewers. The $\Pi$ is a set of all possible  permutations of the reviewers which is $n!$. The corresponding collaboration matrix of an arbitrary ordering $\pi\in\Pi$ is $A^\pi\in \mathbb{R}^{n\times n}$ with $A_{i,j}^{\pi} = \mathbf{1}[\pi(r_i),\pi(r_j)\in E]$, where $E$ is the set of edges. We aim to learn a set of distributions $p(G^{TW}_i), G^{TW}_i\in\{G^{TW}_1, \dots, G^{TW}_i, \dots,G^{TW}_M\}$ for all possible groups where each $G^{TW}_i$ represents a grouping of reviewers within a collaboration matrix.\\
Now, a mapping function $f_S$ is applied to co-review networks to output the sequences, enabling us to use an autoregressive model:
\begin{equation}
\label{eq:adj-seq}
S^{\pi} = f_S(G^{TW},\pi) = (S_1^\pi,S_2^\pi,...,S_n^\pi)
\end{equation}
In Eq. \ref{eq:adj-seq}, $S_i^\pi\in \{0,1\}^{i-1}, i\in\{1,...,n\}$ represents the collaboration of the user $i$ with other users called the ``collaboration vector" ($S_i^\pi = (A_{1,i}^{\pi},...,A_{i-1,i}^{\pi})^T,\forall i \in \{2,...,n\}$) between reviewer $\pi(r_i)$ and previous reviewers $\pi(r_j),j\in \{1,...,i-1\}$ in a group. To fully characterize $p(G^{TW})$, $p(S^\pi)$ is learned:
\begin{equation}
\label{eq:G-joint}
p(G^{TW}) = \sum_{S^\pi}p(S^\pi)\mathbf{1}[f_G(S^\pi) = G^{TW}]
\end{equation}
We rewrite the $p(S^\pi)$ as an autoregressive conditional distribution:
\begin{equation}
\label{eq:autoregressive-col}
    p(S^\pi) = \prod_{i=1}^{n+1}p(S_i^\pi|S_{i}^\pi,S_{i-1}^\pi,...,S_{1}^\pi) = \prod_{i=1}^{n+1}p(S_i^\pi|S_{<i}^\pi)
\end{equation}
where $S_{<i}^\pi = \{S_{i-1}^\pi, S_{i-2}^\pi, \dots, S_{1}^\pi\}$. To fully capture the relations, the HIN-RNN expands Eq.~\ref{eq:autoregressive-col} as below:
\begin{equation}
\label{eq:autoregressive-edge}
p(S_i^\pi|S_{<i}^\pi) = \prod_{j=1}^{i-1}p(S_{i,j}^\pi|v_i, S_{i,<j}^\pi,S_{<i}^\pi)
\end{equation}
where the $v_r$ is the current reviewer's representation, $S_{i,<j}^\pi = \{S_{i,j-1}^\pi, S_{i,j-2}^\pi, \dots, S_{i,1}^\pi\}$, and $S_{i,j}^\pi$ is 1 if there is collaboration between reviewers $i$ and $j$ (the refined collaboration link between $r_2,r_3$ in Fig.~\ref{fig:framework}). The structure of the HIN-RNN model is displayed in Fig.~\ref{fig:HIN-RNN}.\\
To parameterize two autoregressive models, two RNNs are utilized. First:
\begin{equation}
h_i = f_{1}(h_{i-1},S_{i-1}^\pi,v_{i})
\end{equation}
where the $h_i$ encodes the state of the groups (reviewers plus their collaboration matrix) up to the reviewer $i$ and the $f_1$ is the function learned by the RNN. Next, we obtain the collaboration matrix of the current reviewer: 
\begin{equation}
S_i^\pi = f_{2}(h_i)
\end{equation}
where $S_i^\pi$ encodes the collaboration vector obtained from the function $f_2$ using the RNN. 

\subsection{Spatio-Temporal Modeling}
\label{sec:temporal-modeling}
Intuitively, group activities of fraudster groups increases gradually over time (See Figs.~\ref{fig:TWs-yelp} and \ref{fig:TWs-amazon}), the final time windows encode richer temporal characteristics. In other words, as time goes forward a better resolution of the fraudster groups' temporal activities is obtained. Hence, acquiring the collaboration matrix of the final time window provides a more accurate modeling. So, in this step we aim to obtain the collaboration matrix of the groups. So far, we modeled the spatial relations between reviewers, as $p(G^{TW}), \forall TW\in\{1, \dots, N\}$ (See Eq.~\ref{eq:G-joint}), where $N$ is the total number of time windows. Similar to Sec.~\ref{sec:spatio-modeling}, first the subgraphs for different time windows are mapped to a collaboration matrix of $C^{TW}, C\in \mathbb{R}^{M\times n\times n}$ with $C^{TW}_{g, i,j} = \mathbf{1}$, if reviewers $i$ and $j$ form a collaboration in group $g$. To obtain the collaboration matrix of $g$ in $N+1^{th}$ time window, we model the relation of the collaboration matrices in time windows as an autoregressive model:
\begin{equation}
    \label{eq:rnn-decoding}
    p(C_g)  = p(C_g^{1}, ..., C_g^{N+1}) = \prod^{N+1}_{TW=1}p(C_g^{TW}|C_g^{TW-1}, \dots, C_g^1)  
\end{equation}
The output of this step is the corresponding subgraph of the collaboration matrix in $N+1^{th}$ time window. To parameterize the autoregressive model, we employed a simple one-to-many RNN to model the hidden states: 
\begin{equation}
    \label{eq:temporal-hidden}
    h_g^{TW} = \partial(W_hh_g^{TW-1} \oplus W_CC_g^{TW})
\end{equation}
where the $h^{TW}_g$ is the hidden state of the group $g$ at $TW^{th}$ time window, the $W_h$ is the weight matrix of hidden states and the $W_C$ is the weight matrix of the transition matrix. The collaboration matrix of $N+1^{th}$ time window is obtained through:
\begin{equation}
    \label{eq:temporal-final-tw}
    C_g^{N+1} = \tanh(h_g^{N})
\end{equation}
\subsection{Representation Refinement Through Graph Convolutional Network}
\label{sec:gcn-refinement}
\begin{figure}
\centering
\includegraphics[width=\linewidth]{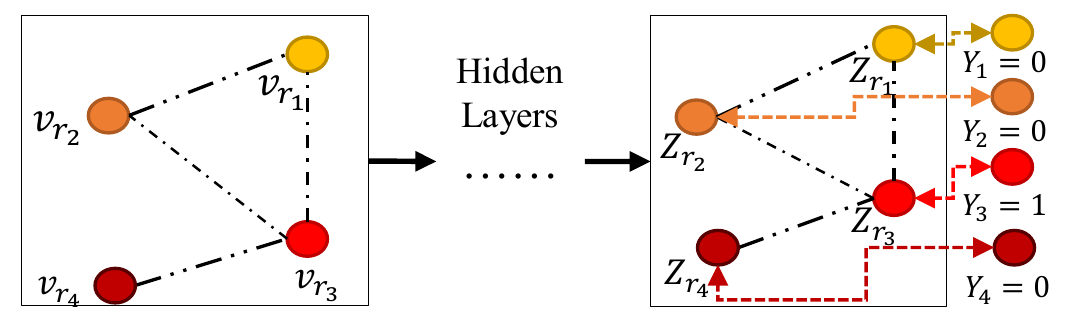}
\caption{ The schematic depiction of a multi-layer Graph Convolutional Network (GCN) for training.} 
\label{fig:GCN}
\end{figure}

Given the refined spatio-temporal relations captured as a collaboration matrix in the previous steps, in this step, we employ the Graph Convolutional Network (GCN) to refine the representation of the reviewers in the collaboration matrix. As such refined representations are trained based on 1) the collaboration matrix and 2) the labels for each reviewer, they bring different merits: \textbf{First,} the reviewers with a similar behavior are represented closer to each other in the feature space. \textbf{Second,} the outlier reviewers (a genuine reviewer in a fraudster group, or a fraudster in a genuine group) are distant from the majority of the reviewers (See Sec. \ref{sec:refinement-effectiveness}) in the feature space.\\
The GCN uses the concept of a layer-wise propagation for neural network models to encode both the adjacency matrix (collaboration matrix) and the labels of the nodes (reviewers). Therefore, the model is defined as $f(V_g,C_g)$, where $V_g$ is a 2D vector representation of the reviewers in the group $g$,  and $C_g$ is the collaboration matrix of group $g$ in the $N+1^{th}$ time window. Conditioning $f(.)$ on the collaboration matrix of each group helps the model to distribute the gradient information from the supervised loss and will enable the GCN to learn representations of the reviewers. With the introduction of a multi-layer neural network to model $f(V_g,C_g)$, efficient information propagation is obtained, while the prediction is labeled on both the representations and the collaboration matrix. Therefore, in this study, we employ a two-layer GCN, with an assumption that the collaboration matrix for each group is symmetric. GCN defines a self-connected adjacency matrix (collaboration) matrix, $\tilde C_g = C_g + I$, where $I$ is the identity matrix. Then a degree matrix is calculated for each node (reviewer) in group $g$ as $\tilde D_{g,i,i} = \sum_j \tilde C_{g,i,j}$. Given the definitions, to obtain the final representation based on the collaboration matrix and representations, the GCN assumes $\hat C_g = \tilde D_g^{-\frac{1}{2}}\tilde C_g\tilde D_g^{-\frac{1}{2}}$. $\hat C_g$ is used to calculate the final representation, $Z_g = \hat C_gV_g\theta$ where $\theta$ is the parameter that GCN learns. The model is then formulated as:
\begin{equation}
    \label{eq:gcn}
    Z_g = f(V_g,C_g) = softmax(\hat C_g.ReLU(\hat C_gV_gW^{(0)})W^{(1)})
\end{equation}
where $W^{(0)}$ is the weight matrix of the input-to-hidden layer and $W^{(1)}$ is the weight matrix of the hidden-to-output layer. To output the label for each review GCN defines $softmax(x_i)= \frac{exp(x_i)}{\sum_i exp(x_i)}$. To train the weight matrices, we define a cross-entropy over the reviewers' labels:
\begin{equation}
    \label{eq:gcn-CE}
    loss = - \sum_{l\in L}\sum^{F}_{f=1} Y_l\ln(Z_f)
\end{equation}
where $F$ is the feature dimension, and $L$ is the set of labels (genuine, fraudster). This ensures further refinement of the reviewers' representation based on the reviewer type. A schematic of GCN is displayed in Fig.~\ref{fig:GCN}.

\begin{algorithm}[hbt!]
 \caption{Proposed Classification Algorithm}
\label{alg:overall-alg}
\textbf{Output:}\
The label of each group\;
\textbf{Input:}\
$T$ review texts of $R$ reviewers and ratings $rate$, written on $I$ items, $N$ is the number of time windows, $M$ number of groups\;
 
 \% \textbf{Step 1:} spatial modeling\;
 \For{$TW \gets 1$ to $N$}
 {
\%  candidate groups in time windows\;
           \uIf{$r_x,r_y$ co-review same item $i\in I, \forall r_x,r_y\in R$}{
            \% Link the possible collaborating reviewers\;
                $E(x,y)$ = 1; 
            }
\%  reviewer representation\;
    \For{$r \gets 1$ to $R$} 
    {
        \% Tokenize $t\in T$  to $S$ sentences\; 
        $\{s_1,s_2,...s_S\} \leftarrow tokenize(t_r)$\; 

        \% Sentence Representation\;
        \For{$s \gets 1$ to $S$} 
        {
            \% Tokenize $S_s$ to $n$ words\;
            $\{w_1, \dots, w_n\} \leftarrow tokenize(S_s)$\; 
            \% Word embeddings\;
            $\{e_{w_1}, \dots, e_{w_n}\} \leftarrow WE(\{w_1, \dots, w_n\})$\;
            \% The sentence representation\;
            $e_{s} \leftarrow average(\{e_{w_1}, \dots, e_{w_n}\})$\; 
        }
        \% Negative Ratio of reviewer $r$\;
        $NR_r\leftarrow NR(rate_1,rate_2,...,rate_m)$\;
        \% Final representation of $r$\;
        $v_r \leftarrow max(concat({e_{s},\forall s\in S})) \oplus NR_r$\; 
    }
 \% spatial modeling through HIN-RNN\; 
    \For{$g \gets 1$ to $M$}
    {
    \For{$i \gets 1$ to $R$}
    {
        $h_{r_i} \leftarrow RNN_1(h_{r_{i-1}},E(0:i-1,0:i-1),v_i)$\;
        \% Collaboration matrix of reviewer $i$\;
        $C^{TW}_{g,i} \leftarrow  RNN_2(h_{r_i})$; 
    }
    }
}
\% \textbf{Step 2:} temporal modeling\; 
     \For{$g \gets 1$ to $M$}
    {
\For{$TW \gets 1$ to $N$}
 {
 $h_g^{TW} \leftarrow \partial(W_hh_g^{TW-1}\oplus W_CC_g^{TW})$
 }
 $C_g^{N+1} \gets \tanh{h_g^{N}}$\;
 $C_g \gets C_g^{N+1}$\;
 
}
\% \textbf{Step 3:} GCN refinement\; 
     \For{$g \gets 1$ to $M$}
    {
 \For{$r \gets 1$ to $R$}
 {
 $V_g \leftarrow v_1 \oplus \dots \oplus v_r$\;
 }
 $Z_g \leftarrow softmax(\hat C_g.ReLU(\hat C_gV_gW^{(0)})W^{(1)})$\;
}
\% \textbf{Step 4:} group classification\; 
     \For{$g \gets 1$ to $M$}
    {

        $g \leftarrow g - $\{reviewers in the cluster with the minimum reviewers based on $Z_g$\} \;
        $v_g\leftarrow mean(v_r)$ for $r\in g$\;
        $label(g)\leftarrow fc(v_g)$\;
}
\end{algorithm}

\subsection{Group Classification}
\label{sec:classification}
Our proposed group classification method follows three simple steps: clustering to remove the outlier reviewers, group-level representation using SoWE, and a fully-connected layer. 
\subsubsection{Clustering}
\label{sec:clustering}
In this step, 
we first use a simple K-means algorithm to cluster the reviewers in a group based on their representation. Intuitively, the reviewers in a group can be categorized at most into two clusters (a mix of genuine and fraudster reviewers). To determine the possibility of outliers' existence in groups, we use the Total Sum of Square (TSS), as the sum of squared deviations from the overall mean, Within Sum of Square (WSS), as the sum of the squared deviations within a cluster, and Between Sum of Square (BSS), as the sum of the squared deviations between the clusters. 
For TSS we calculate the element-wise squared between each reviewer's representation ($Z$) and the centroid:
\begin{equation}
    \label{eq:1-NN}
    TSS_i = \sum^{n_c}_{j=1}(Z_{i,j} - c_{i})^2
\end{equation}
where $Z_{i,j}$ is the value of the $i^{th}$ element of the $j^{th}$ reviewer representation (with $1\leq i\leq 100$, being the representation vector dimension), $n_c$ is the number of reviewers in a group and $c_{i}$ is the $i^{th}$ element from the obtained centroid for the group in the clustering algorithm. \\
For the WSS we calculate the element-wise squared for each reviewer's representation in a cluster and the corresponding centroid:
\begin{equation}
    \label{eq:2-NN}
    WSS_i = \sum^{2}_{m=1}\sum^{n_{c_m}}_{j=1}(Z^{(m)}_{i,j} - c^{(m)}_{i})^2
\end{equation}
where $n_{c_m}$ is the number of reviewers in cluster $m$ and $Z^{(m)}_{i,j}$ is the value of the $i^{th}$ element of the $j^{th}$ reviewer in the $m^{th}$ cluster. \\
To obtain the final distance value we first calculate the element-wise BSS of the representation:
\begin{equation}
    \label{eq:BSS}
    BSS_i = TSS_i - WSS_i
\end{equation}
Next, we calculate the second-norm of $BSS_i$:
\begin{equation}
    \label{eq:d1-2nd-norm}
    |BSS| = \frac{1}{n_c}\sqrt{\sum_i BSS_i}
\end{equation}
Then if $BSS<0.5$, the group is considered as a group with no outliers. Otherwise, the group is considered as a mix group, and then the cluster with more reviewers is regarded as dominant cluster and is selected for the next step.  
\subsubsection{Fully-Connected Layer}
\label{sec:fc-layer}
In this step, we first calculate the SoWE for each group through an element-wise average over the representations of the selected cluster from the previous step (Sec.~\ref{sec:clustering}) to obtain a final representation of each group. 
Finally, a fully connected layer is trained based on the obtained representation.
The proposed approach is presented in Algorithm \ref{alg:overall-alg}.

\section{Experimental Evaluation}
\label{sec:experimental-evaluation}
To demonstrate the effectiveness of our proposed approach, we compare our approach to the most recent approaches reviewed in Sec.~\ref{sec:related-works}: Zhang \textit{et al.}~\cite{Zhang2020}, Ji \textit{et al.}~\cite{JI2020454}, and the HIN-RNN by Shehnepoor \textit{et al.}~\cite{shehnepoor2021hinrnn}. 
\subsection{Experimental Setup}
\label{sec:epxerimental-setup}
We trained a 100-dimension Continuous Bag of Words (CBoW) embeddings on available datasets (Yelp and Amazon) with a window size of 2 and batch size of 512, as CBoW has shown to be effective in fraud review detection~\cite{Zhang2016,dfraud,shehnepoor2021hinrnn}. To train the HIN-RNN, we used two RNNs in the spatial modeling step with Gated Recurrent Units (GRUs), one learns the hidden state of the group with a hidden state size of 128, and the other learns the collaboration matrix with a hidden state size of 16. Two RNNs were trained jointly with a learning rate of 0.003 and in 3000 epochs to predict the co-review relations between reviewers. To model the temporal behavior, we used an RNN with $10^{-4}$ as the learning rate and binary cross entropy as the objective function. For the training of the GCN, the learning rate was $10^{-5}$, the training epochs were 100, and the cross-entropy was used as the objective function.

\subsection{Datasets}
\label{sec:datasets}
\begin{center}
\begin{table}[hbt!]
\centering
\caption{List of datasets used in our current study.}\label{tab:datasets}
\begin{tabular}{|c|c|c|c|c|}
\hline
Dataset &  Reviewers & Items & Reviews & Candidate Groups\\
\hline
\hline
Yelp &  260,277 & 5,044 & 608,598 & 9,952\\
Amazon &  42,655 & 6,822 & 53,777 & 2,194\\
\hline
\end{tabular}
\end{table}
\begin{center}
\begin{table*}
\centering
\caption{A comparison between the results on the proposed approach and three state-of-the-art studies \cite{Zhang2020,JI2020454,shehnepoor2021hinrnn}}\label{tab:comparison}
\begin{tabular}{|c|l|c|c|c|c|c|c|}
\hline
\multicolumn{2}{|l|}{Datasets} & \multicolumn{3}{c|}{Yelp} & \multicolumn{3}{c|}{Amazon} \\
\hline
\hline
\multicolumn{2}{|l|}{Metric} &  Precision & Recall & F1-value & Precision & Recall & F1-value\\
\hline
\multirow{3}{*}{Existing Approaches} &Zhang \textit{et al.} \cite{Zhang2020}&  0.70 & 0.20 & 0.32 & 0.80 & 0.45 & 0.58 \\
&Ji \textit{et al.} \cite{JI2020454} &  \textbf{0.83} & 0.60 & 0.69 & 0.82 & 0.92 & 0.86 \\
&Shehnepoor \textit{et al.}~\cite{shehnepoor2021hinrnn} & 0.81 & \textbf{0.82} & \textbf{0.81} & \textbf{0.85} & \textbf{0.90} & \textbf{0.87} \\ \hline
\multicolumn{2}{|l|}{spatial modeling} & 0.74 & 0.70 & 0.72 & 0.85 & 0.90 & 0.87 \\
\multicolumn{2}{|l|}{spatial modeling + Temporal Modeling} & 0.76 & 0.80 & 0.78 & 0.83 & 0.85 & 0.84 \\
\multicolumn{2}{|l|}{spatial modeling + Temporal Modeling +  GCN} & 0.83 & 0.93 & 0.86 & 0.87 & 0.93 & 0.90 \\
\multicolumn{2}{|l|}{spatial modeling + Temporal Modeling +  GCN + Clustering} & \textbf{0.86} & \textbf{0.94} & \textbf{0.93} & \textbf{0.89} & \textbf{0.95} & \textbf{0.92} \\
\hline
\end{tabular}
\end{table*}
\end{center}

\end{center}
Similar to the previous studies~\cite{JI2020454,Zhang2020}, we used the Yelp dataset and the Amazon dataset to demonstrate the scalability of the proposed approach. The Yelp dataset contains reviews ranging from ``20-Oct-2004" to ``10-Jan-2015" alongside the labels, reviewer id, item id, the rating, and the date of the review. The Amazon dataset provides the same metadata with reviews ranging from ``01-Feb-2000" to ``10-Oct-2010". Fraudster groups are determined as described in Sec.~\ref{sec:group-construction}. The details on datasets are provided in Table \ref{tab:datasets}. For evaluation, we used 80\% of the data for training and 20\% for testing.


\subsection{Main Results}
\label{sec:main-results}
We used three well-known metrics to evaluate the performance of the proposed approach. First, precision:
\begin{equation}
    \label{eq:precision}
    Precision = \frac{TP}{TP + FP}
\end{equation}
where $TP$ is the number of true positive samples and $FP$ is the number of false positive samples. We also use recall:
\begin{equation}
    \label{eq:recall}
    Recall = \frac{TP}{TP + FN}
\end{equation}
where $FN$ is the number of false negative samples. Finally, we also use the F1-value:
\begin{equation}
    \label{eq:f1-value}
    F1-value = \frac{2\times precision\times recall}{precision + recall}
\end{equation}
\subsubsection{Comparison with baselines}
\label{sec:comparison}
We compared the proposed approach against three state-of-the-art approaches. The results are displayed in Table~\ref{tab:comparison}. 
We devised different configurations to show the effectiveness of each step in this study: \\
\textbf{spatial modeling:} In this configuration, we obtain the collaboration matrix of a static network (without temporal modeling). 
Next, the average representation of the reviewers in the group is used to calculate the final label for each group.\\
\textbf{spatial modeling + temporal modeling:} In this configuration, after the spatio modeling step, the collaboration matrix of the $N+1^{th}$ time window is obtained. Similar to the previous configuration, we then calculate the average of the reviewers' representation of the group to predict the label of each group.\\
\textbf{spatial modeling + temporal modeling + GCN:} Given the spatio-temporal modeling of the groups, in this configuration, the GCN is employed to refine the representation of the reviewers. The refined representation of the reviewers is used to predict the group's label.\\
\textbf{spatial modeling + temporal modeling + GCN + clustering:} The whole framework is used in this configuration.\\
As shown in Table~\ref{tab:comparison}, the proposed approach outperforms markedly the most recent state-of-the-art approach (Shehnepoor \textit{et al.}~\cite{shehnepoor2021hinrnn}) by 12\% for recall and 9\% for F1-value on the Yelp dataset. Additionally, the proposed approach improves the performance by 5\% for all metrics on the Amazon dataset. Previous studies overlooked the temporal clues which are effective in capturing the true behavior of reviewers in social media as group fraudsters change their behavior instantly to camouflage. However, the approach by Shehnepoor \textit{et al.}~\cite{shehnepoor2021hinrnn} is still superior to our proposed approach with the spatial modeling alone, and the spatial modeling + temporal modeling. The main reason is that in such configurations, i.e., spatial and spatial + temporal modeling, outlier reviewers' removal is not considered. On the other hand, Shehnepoor \textit{et al.}~\cite{shehnepoor2021hinrnn} removed the outlier reviewers based on the minimum connections, resulting in better performance. The GCN refinement improved the spatio-temporal relations captured in the reviewers' representation. Previous studies also suffered from a limitation in their inabilities to exclude genuine reviewers from a fraudster group or a fraudster from a genuine group. With the representations refined through GCN, we can exclude the outlier reviewers. Outlier reviewers are known to be responsible for the increase in $FN$ and $FP$. As shown in Table~\ref{tab:comparison}, the proposed approach outperforms all previous approaches. 
\begin{figure*}
\begin{subfigure}{0.46\textwidth}
\centering
\includegraphics[width=\textwidth]{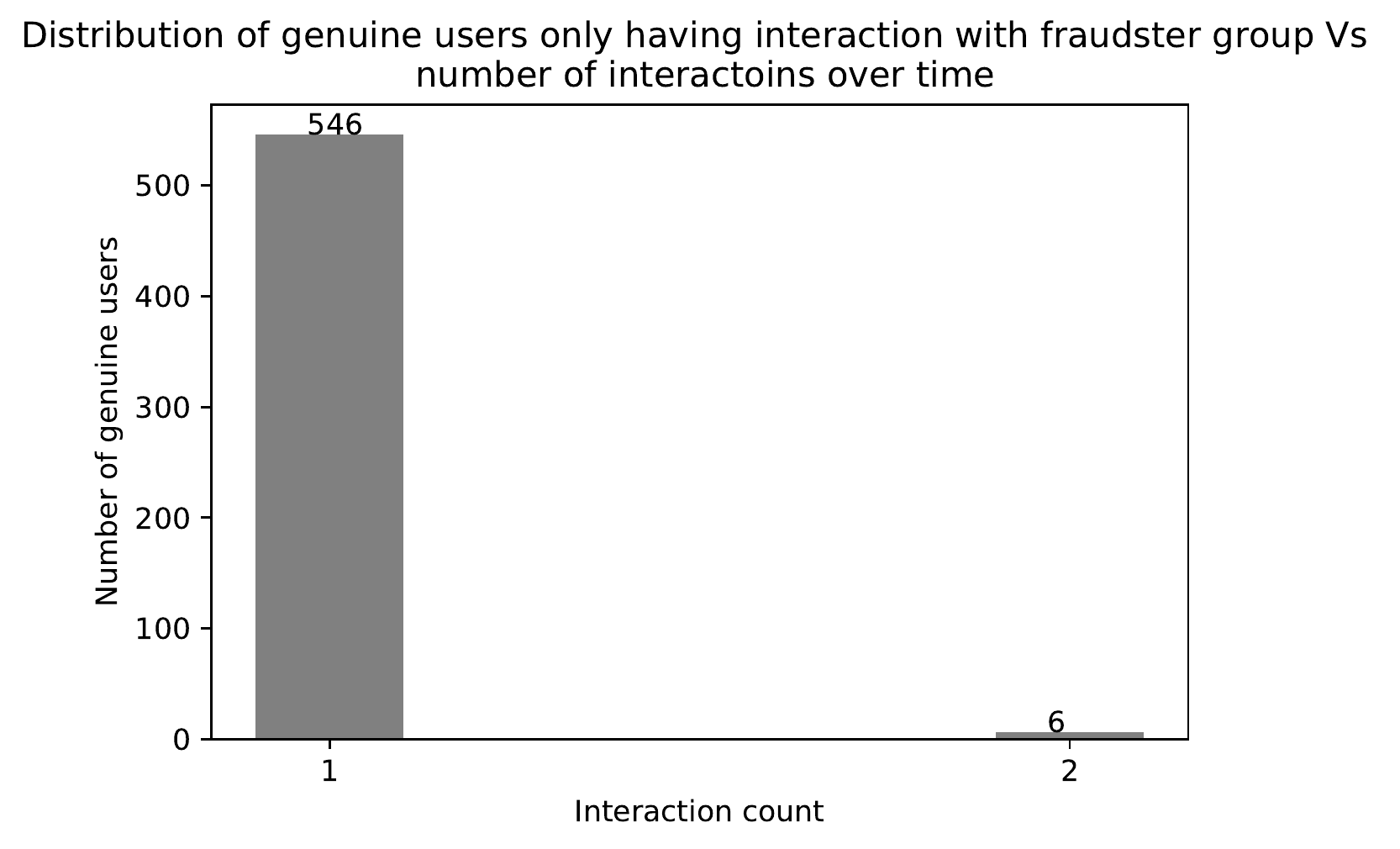}
\caption{Number of genuine reviewers' interactions with fraudster groups in the Yelp dataset.} 
\label{fig:genuine_yelp}
\end{subfigure}
\hfill
\begin{subfigure}{0.46\textwidth}
\centering
\includegraphics[width=\textwidth]{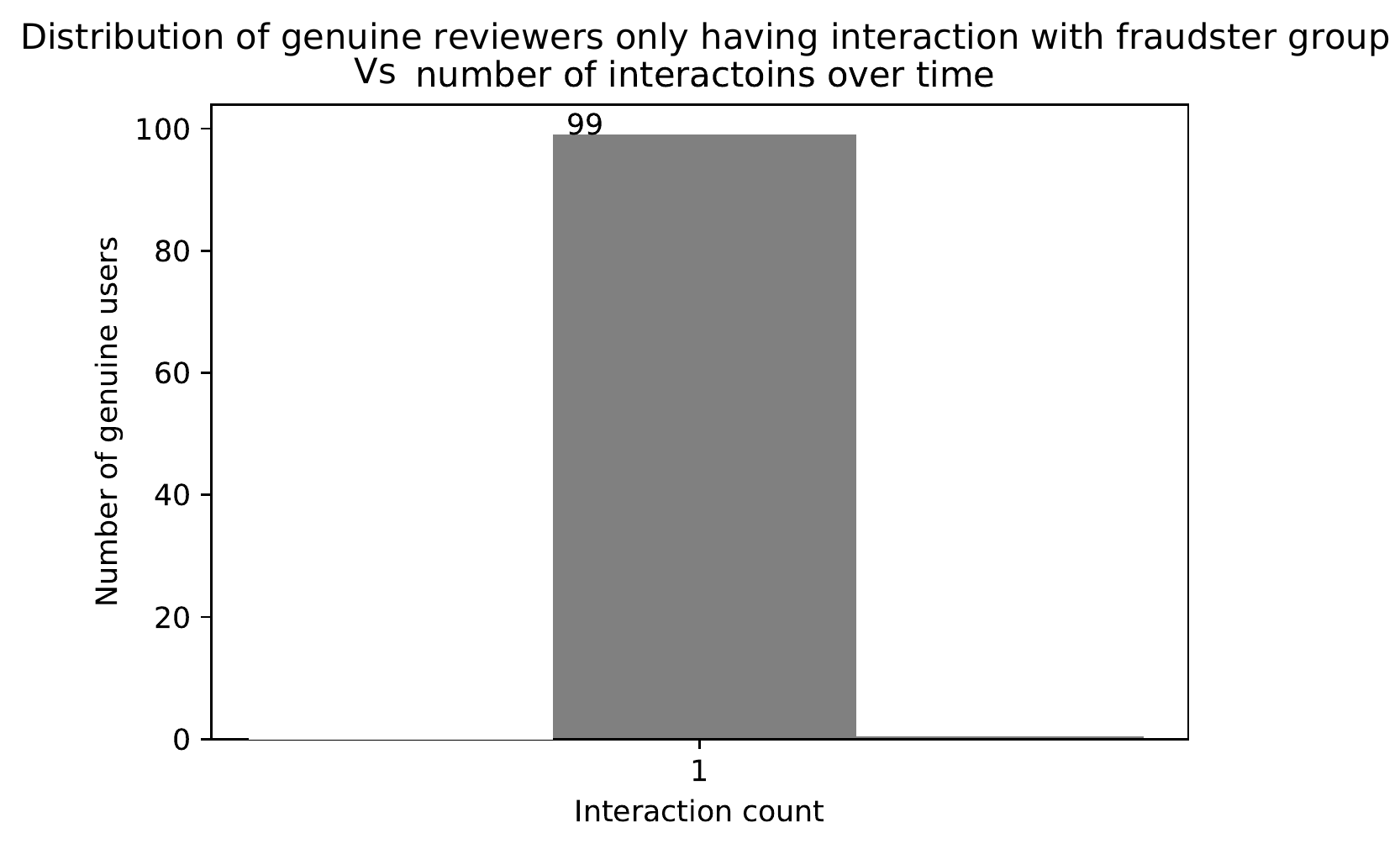}
\caption{Number of genuine reviewers' interactions with fraudster groups in the Amazon dataset.} 
\label{fig:genuine_amazon}
\end{subfigure}
\hfill
\begin{subfigure}{0.46\textwidth}
\centering
\includegraphics[width=\textwidth]{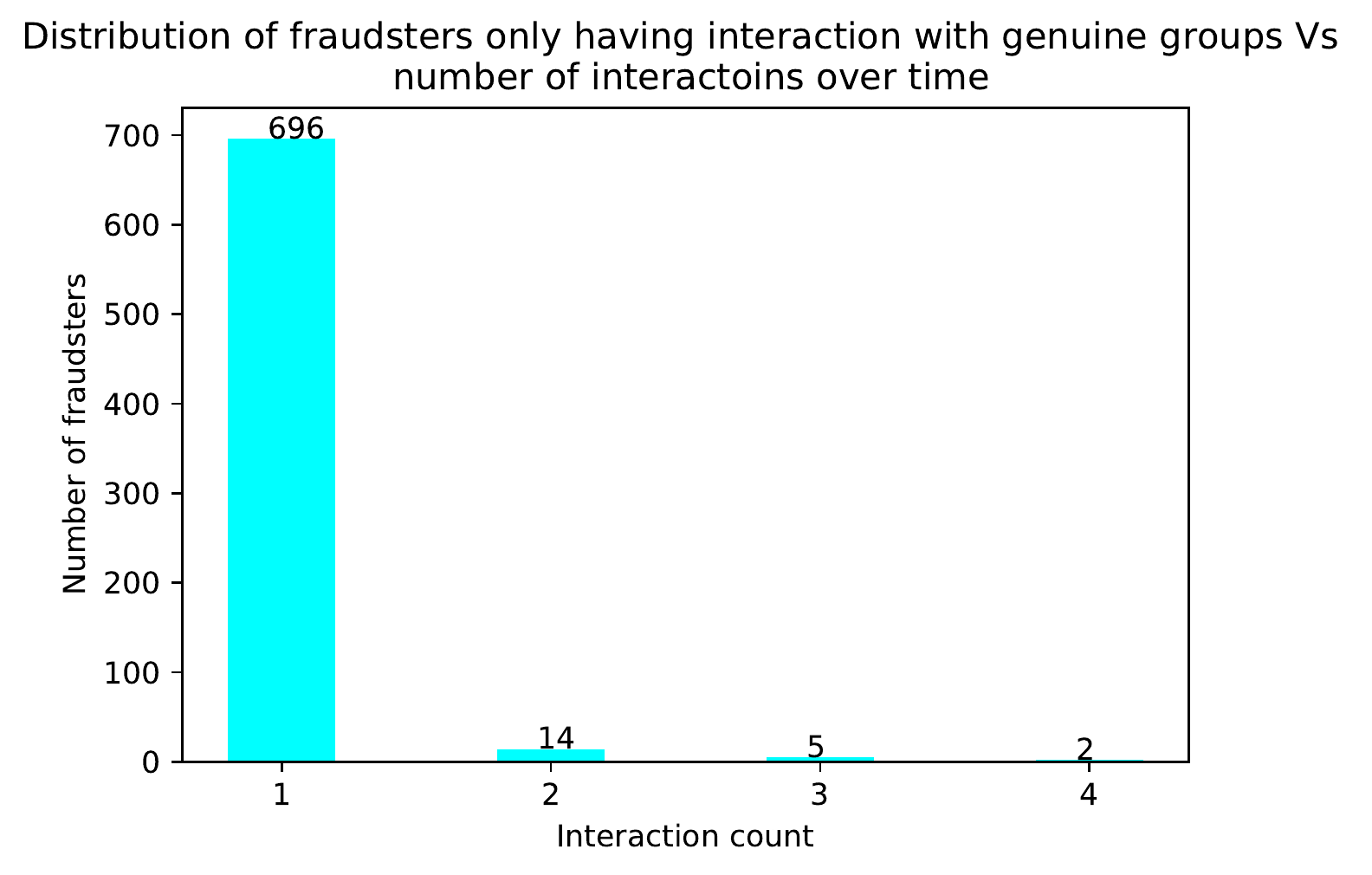}
\caption{Number of fraudster reviewers' interactions with genuine groups in the Yelp dataset.} 
\label{fig:fraudster_yelp}
\end{subfigure}
\hfill
\begin{subfigure}{0.46\textwidth}
\centering
\includegraphics[width=\textwidth]{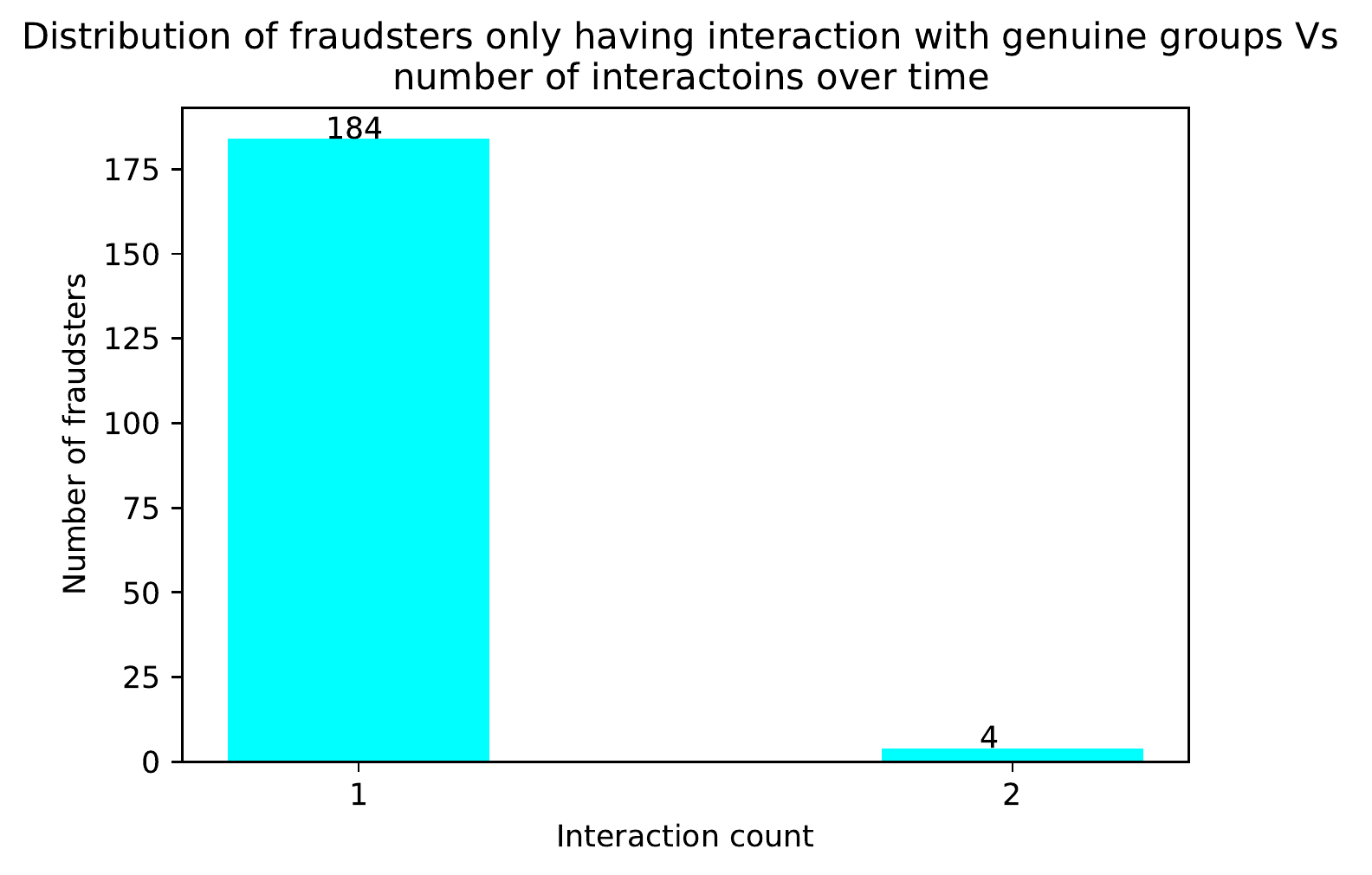}
\caption{Number of fraudster reviewers' interactions with genuine groups in the Amazon dataset.} 
\label{fig:fraudster_amazon}
\end{subfigure}
\caption{Number of interactions between  genuine (fraudster) reviewers in a fraudster (genuine) groups over different time windows.}
\label{fig:temporal-analysis}
\end{figure*}

Intuitively, as the reviewers' representation is refined, the purpose of adding the clustering step is only to achieve a more uniform representation (i.e., a representation of the group with all reviewers of the same type, namely genuine or fraudster). 
As a result, adding the clustering step does not necessarily improve the performance any further.

\subsubsection{Temporal Modeling Effectiveness}
\label{sec:temporal-modeling-effectiveness}
\begin{figure}[ht]
\centering
\includegraphics[width=0.47\textwidth]{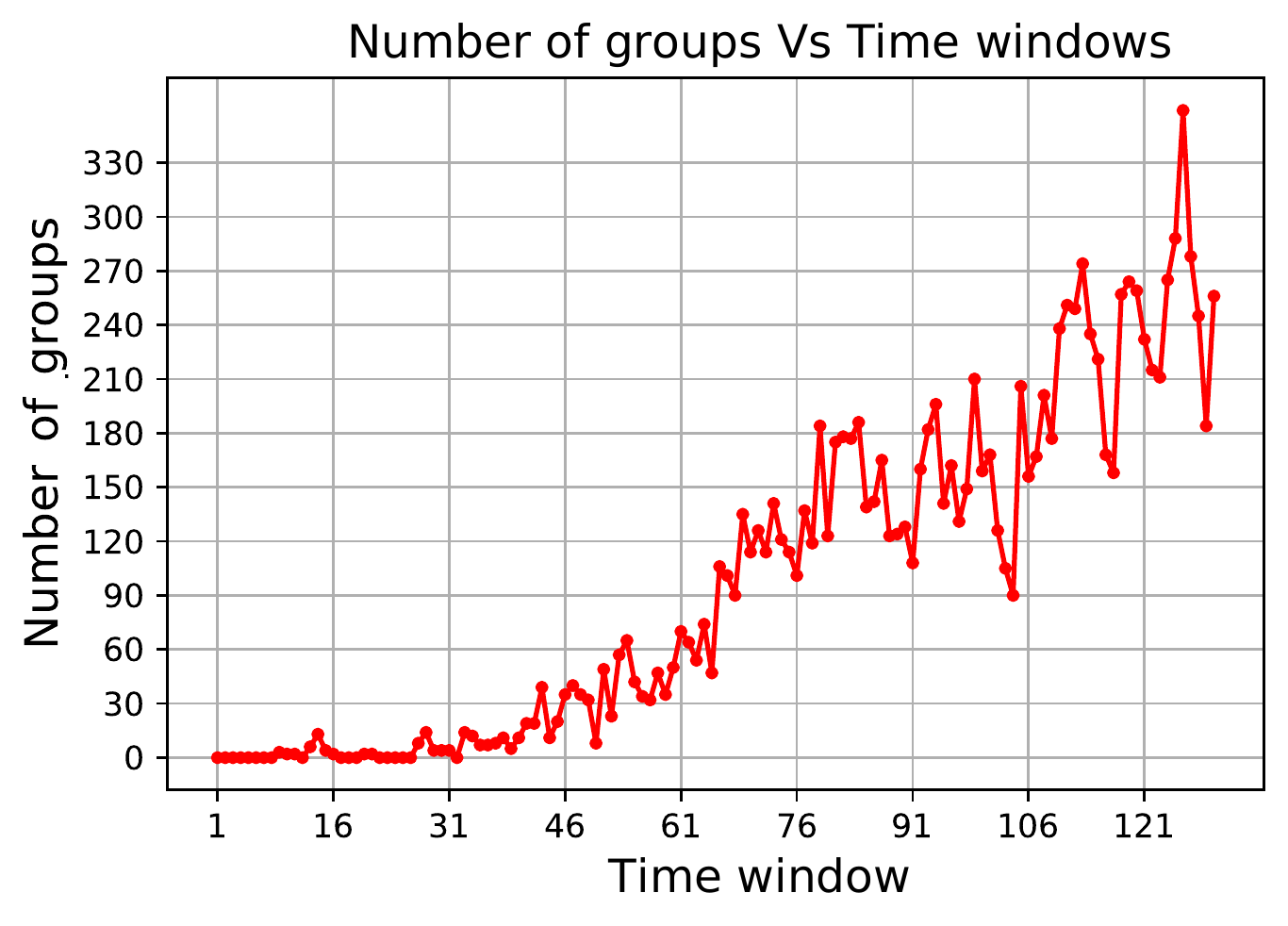}
\caption{The number of groups in different time windows on the Yelp dataset.} 
\label{fig:TWs-yelp}
\end{figure}
\begin{figure}[ht]
\centering
\includegraphics[width=0.47\textwidth]{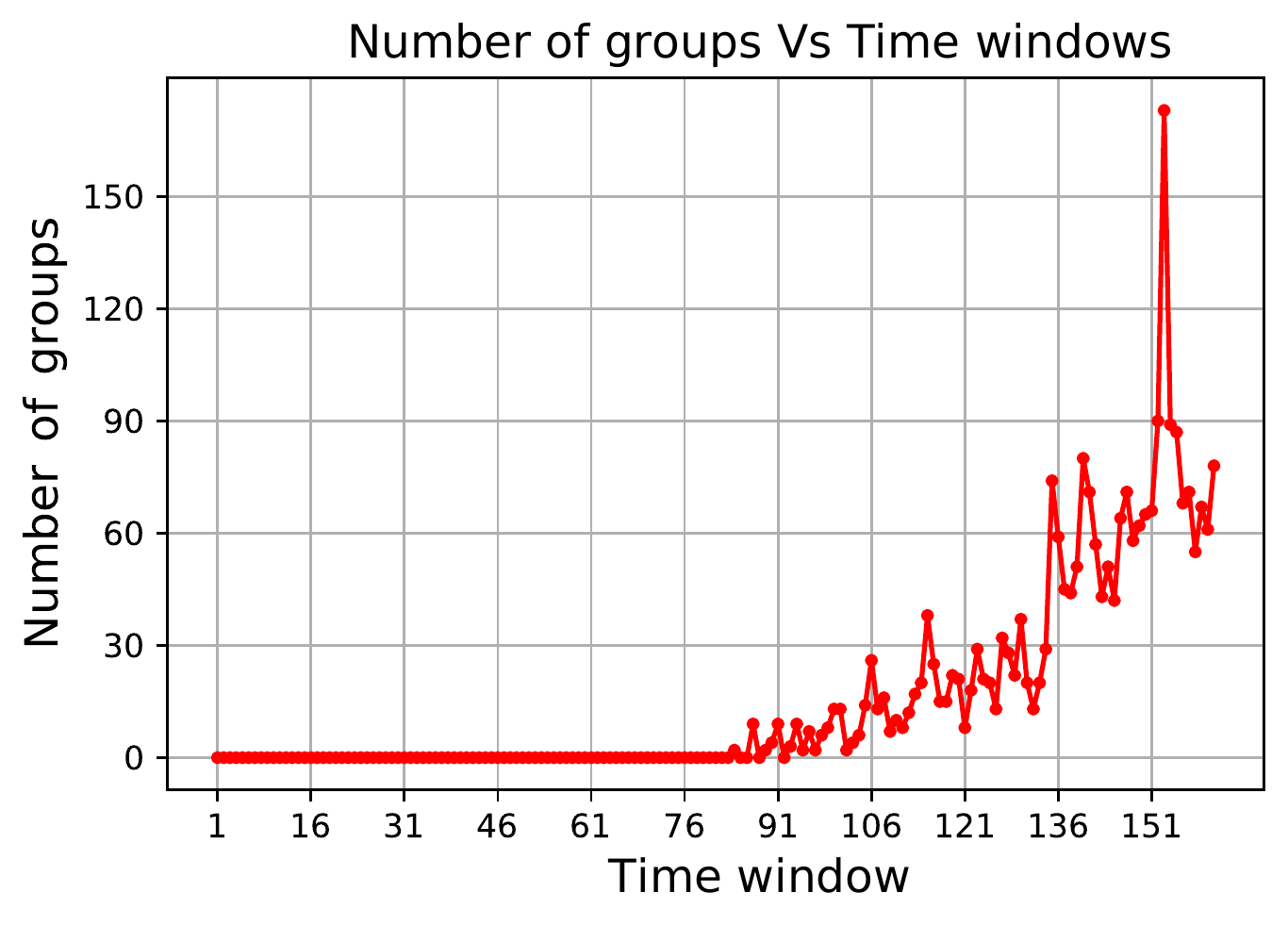}
\caption{The number of groups in different time windows on the Amazon dataset.} 
\label{fig:TWs-amazon}
\end{figure}

In this section, 
we first explain the ablative study in Table~\ref{tab:comparison}, and then we provide
explanations related to temporal trends in the datasets.\\ 
\textbf{Ablative Study:} As shown in Table~\ref{tab:comparison}, the temporal modeling successfully improves the performance of the proposed approach on the Yelp dataset for all three metrics, compared to the spatial modeling. On the other hand, the performance are reduced for the Amazon dataset. This is mostly because the Amazon dataset is considerably smaller than the Yelp dataset. Additionally, the Amazon dataset does not contain groups for almost the first half of the time windows (Fig.~\ref{fig:TWs-amazon}), while in the Yelp dataset (Fig.~\ref{fig:TWs-yelp}) groups are available from the first time window. As such, the Yelp dataset provides more information on the relations between reviewers in a group to the RNN. Given the long-range temporal activity of the reviewers in a group, an RNN improves the performance of the proposed approach, while the lack of temporal activities of reviewers from different groups results in a decrease in the performance on the Amazon dataset.\\
\begin{figure*}
\begin{subfigure}{0.46\textwidth}
\centering
\includegraphics[width=\textwidth]{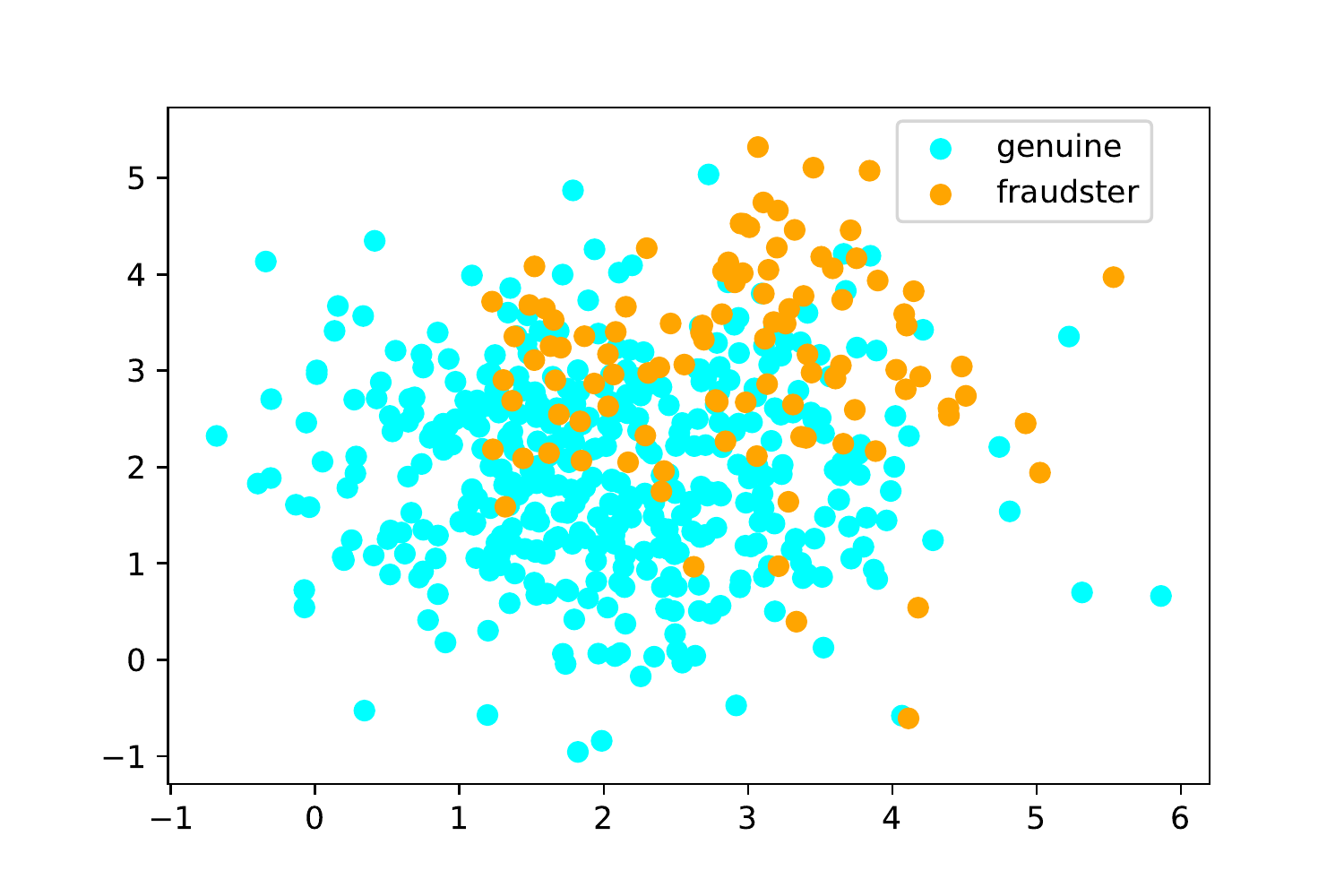}
\caption{A 2D representation of the reviewers in a batch with 512 samples before refinement on the Yelp dataset.} 
\label{fig:yelp-before}
\end{subfigure}
\hfill
\begin{subfigure}{0.46\textwidth}
\centering
\includegraphics[width=\textwidth]{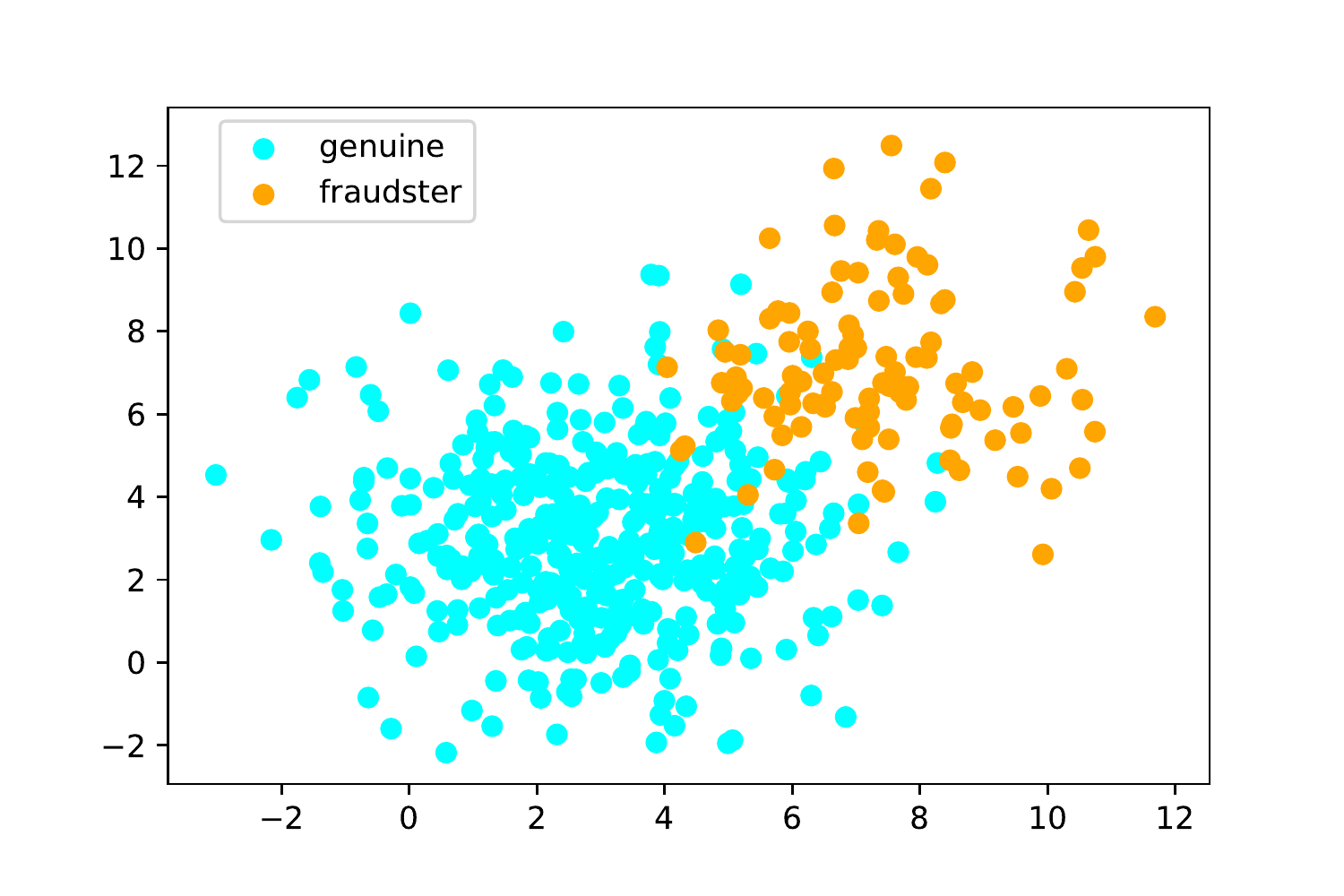}
\caption{A 2D representation of the reviewers in a batch with 512 samples after refinement on the Yelp dataset.} 
\label{fig:yelp-after}
\end{subfigure}
\hfill
\begin{subfigure}{0.46\textwidth}
\centering
\includegraphics[width=\textwidth]{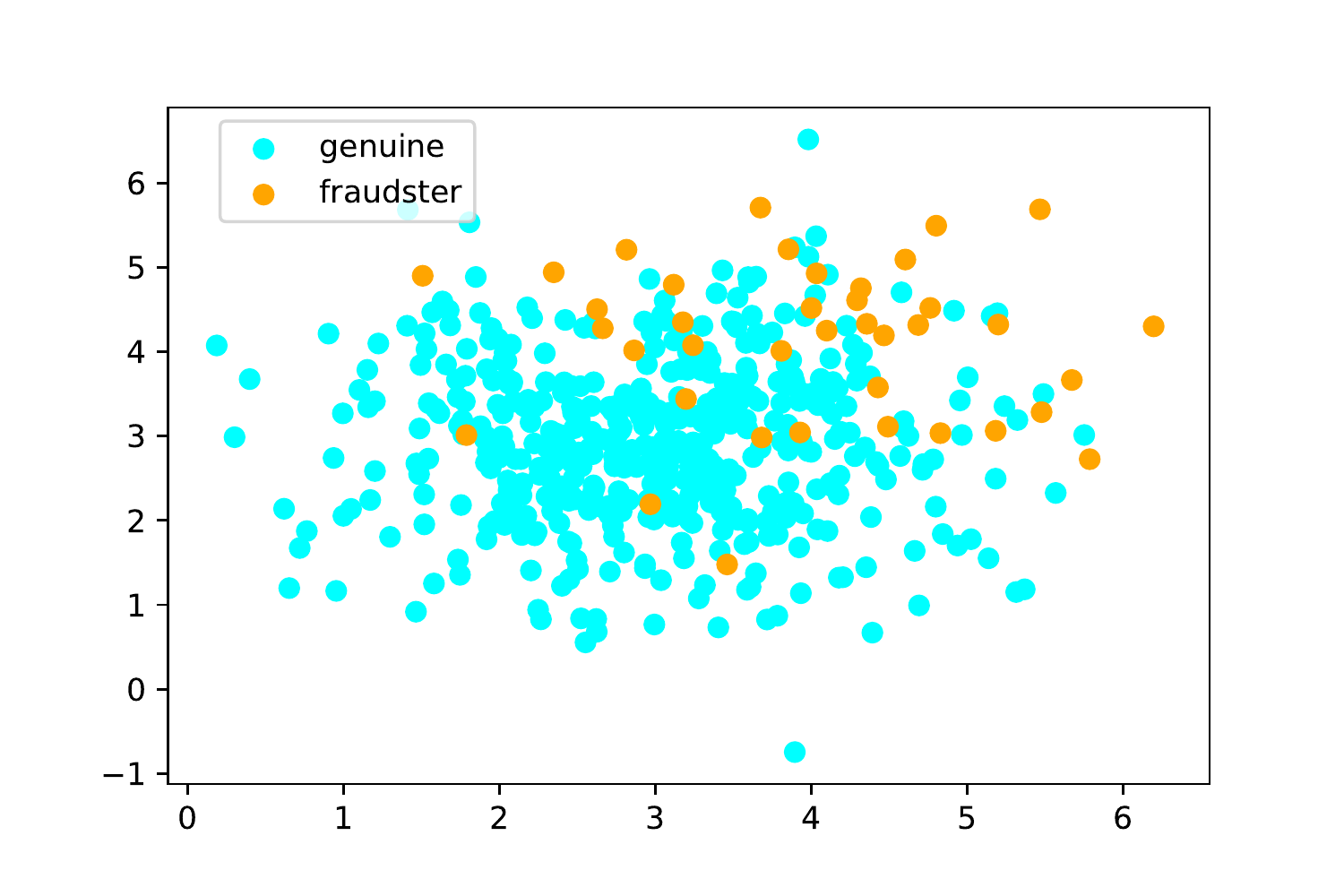}
\caption{A 2D representation of the reviewers in a batch with 512 samples before refinement on the Amazon dataset.} 
\label{fig:amazon-before}
\end{subfigure}
\hfill
\begin{subfigure}{0.46\textwidth}
\centering
\includegraphics[width=\textwidth]{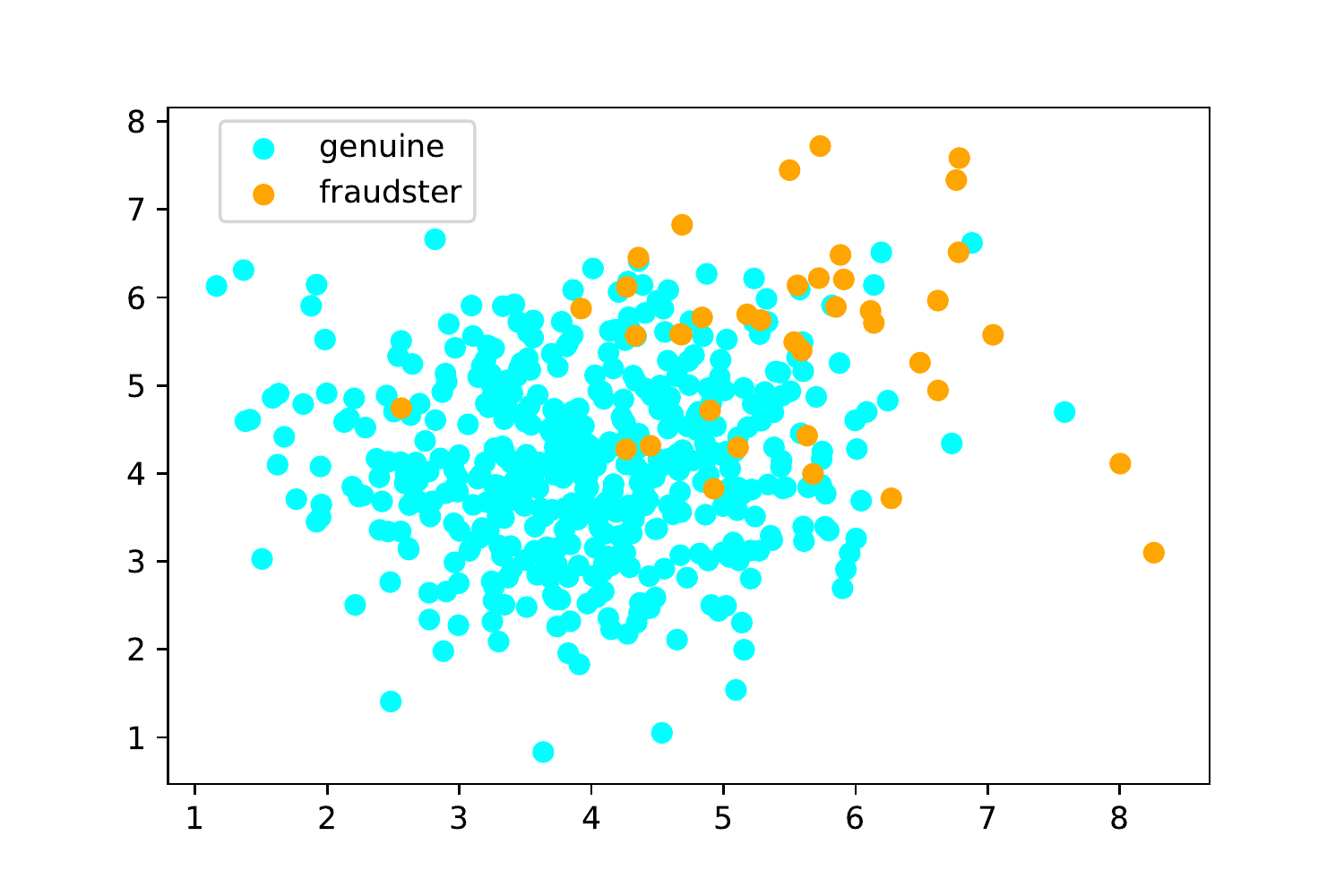}
\caption{A 2D representation of the reviewers in a batch with 512 samples after refinement on the Amazon dataset.} 
\label{fig:amazon-after}
\end{subfigure}
\caption{Refinement effectiveness on reviewers representation on both datasets.}
\label{fig:refinement-effect}
\end{figure*}

\textbf{Observation Analysis:} 
Fig.~\ref{fig:temporal-analysis} shows the number of interactions between genuine (fraudster) reviewers with fraudster (genuine) groups over different time windows. Generally speaking, the genuine reviewers are much less involved in a fraudster group activity in terms of both the number of interactions and also in terms of interaction times, while the fraudsters made multiple contacts with genuine groups over time to camouflage, thus escaping the detection algorithms. Specifically, 21 fraudsters in the Yelp dataset made contacts at least twice with a group of people writing genuine reviews (Fig.~\ref{fig:fraudster_yelp}). In other words, single fraudsters camouflage over time by writing a couple of genuine reviews in collaboration with genuine groups. This helps fraudsters to stay undetected against their fraudster activities afterward. Genuine reviewers on the Yelp dataset are also involved in fraudster groups. As Fig.~\ref{fig:genuine_yelp} shows, 600 different genuine reviewers collaborated with fraudster groups at least once in the Yelp dataset. Such a collaboration is unintentional or is motivated by the members of a fraudster group by highlighting the same interest. As explained in the previous section, reviewers in the Amazon dataset participated in the group activity much less than the reviewers in the Yelp dataset. As such, the number of interactions between genuine (fraudster) reviewers and fraudster (genuine) groups decreases, accordingly. However, 188 fraudsters were involved in a genuine activity to camouflage. Interestingly, genuine reviewers made contact only once with fraudster groups. 
\subsubsection{Effectiveness of refinement}
\label{sec:refinement-effectiveness}
In this study, we employed the GCN to refine the representations based on spatio-temporal relations captured between reviewers. To demonstrate the effectiveness of the GCN refinement, we first provide experimental results 
from the ablative study (Table~\ref{tab:comparison}), and then analyze the observations to analyze how the GCN refines the reviewers' representation on two datasets.\\ 
\textbf{Ablative Study:} As shown in Table~\ref{tab:comparison}, the GCN refinement significantly improves the performance of the proposed approach. The GCN refinement is the most effective step to improve the performance of the proposed approach, as an improvement of 5-10\% on the Yelp dataset for three metrics is obtained. Similarly, the improvement on the Amazon dataset is 4-8\%. To refine the representations, in the training phase, the GCN takes the representations and the collaboration matrix and fine-tunes the parameters based on the reviewers' labels. The main reason behind such a performance gain is because GCN is able to consider both labels and the collaboration matrix in the refinement. This helps the proposed approach better utilisation of the labels of outlier reviewers in a group 
thus preventing fraudsters from escaping detection, and genuine reviewers being mistaken as fraudsters.\\
\textbf{Comparative Analysis of Refinement:} 
To observe how the GCN refinement affects the reviewers' representation, we devised an experiment which first transforms the reviewers' representation from a 100-dimensional feature space to a 2-dimensional feature space using Principal Component Analysis.
For this purpose, we used test batches of 512 reviewers with the maximum number of fraudsters in each dataset (102 fraudsters in the test batch from the Yelp dataset, and 39 fraudsters in the test batch from the Amazon dataset). After the transformation, the two most significant dimensions are considered to represent the feature space. Fig.~\ref{fig:refinement-effect} shows the effectiveness of the GCN refinement on representations of the reviewers.\\
As shown, the GCN effectively refines the genuine and fraudster representation based on the relations in the group on the Yelp dataset. With the refinement, the reviewers are significantly discriminated based on their label (Fig.~\ref{fig:yelp-before} Vs. Fig.~\ref{fig:yelp-after}). Similarly, the reviewers' representation is also refined after applying the GCN to their representation on the Amazon dataset (Fig.~\ref{fig:amazon-before} vs. Fig.~\ref{fig:amazon-after}). However, the refinement on the Amazon dataset is not as discriminative as on the Yelp dataset. Intuitively, much fewer outlier reviewers (and generally genuine/fraudster groups) exist in the Amazon dataset. As a result of fewer contacts by outlier reviewers in a group, the refinement effectiveness also reduces. This results in a lower discrimination on the Amazon dataset, accordingly. 
\subsubsection{Effects of clustering in determining outlier reviewers}
\label{sec:reviewer-removal-effect}
As explained in Sec.~\ref{sec:clustering}, to reduce the effects of genuine (fraudster) reviewer(s) in a fraudster (genuine) group, we used the K-means clustering algorithm to remove the outlier reviewers.\\ 
\begin{figure}[ht]
\centering
\includegraphics[width=0.5\textwidth]{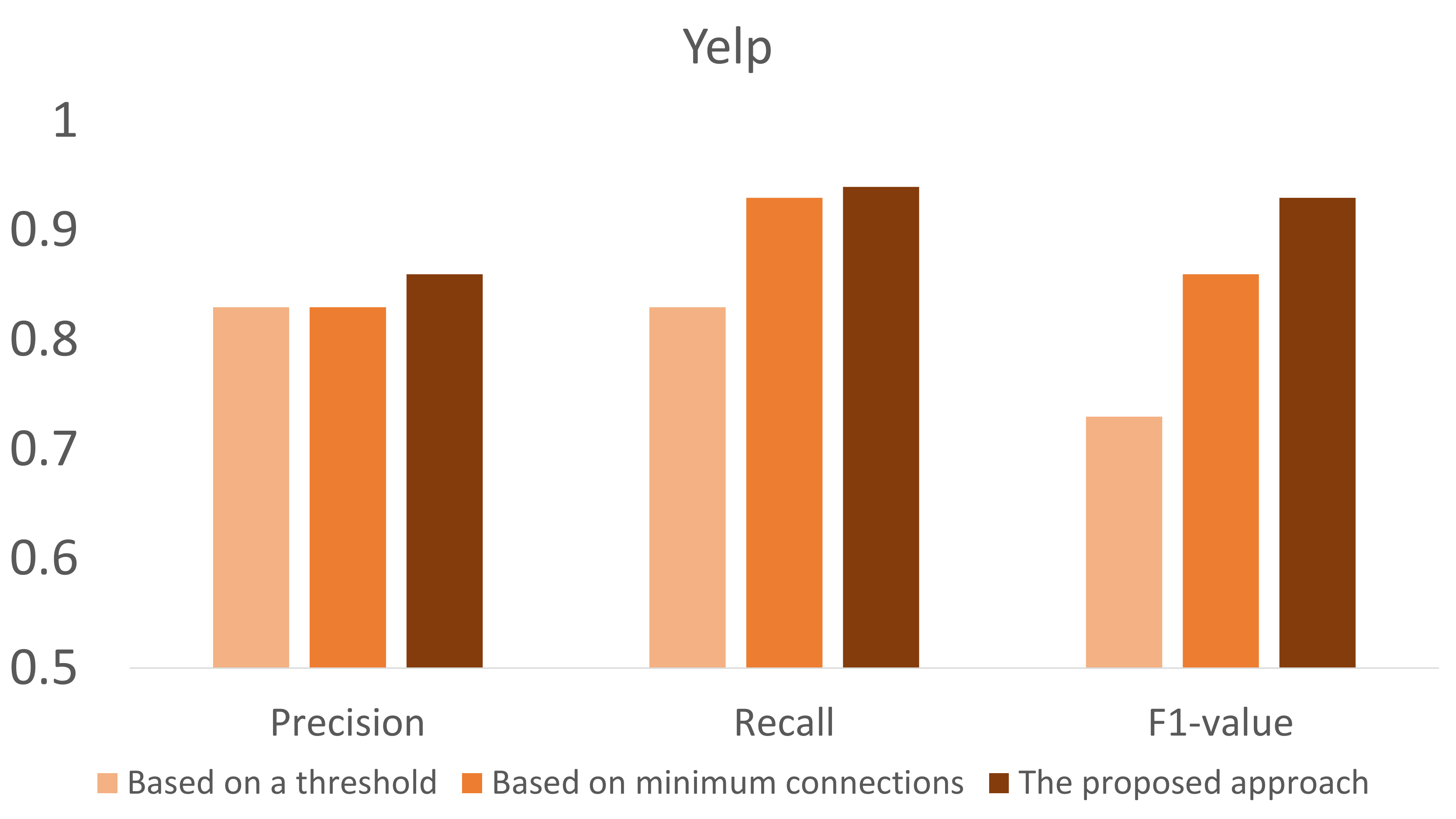}
\caption{The performance of the proposed approach for different removal strategy on the Yelp dataset.} 
\label{fig:clus-str-yelp}
\end{figure}
\begin{figure}[ht]
\centering
\includegraphics[width=0.5\textwidth]{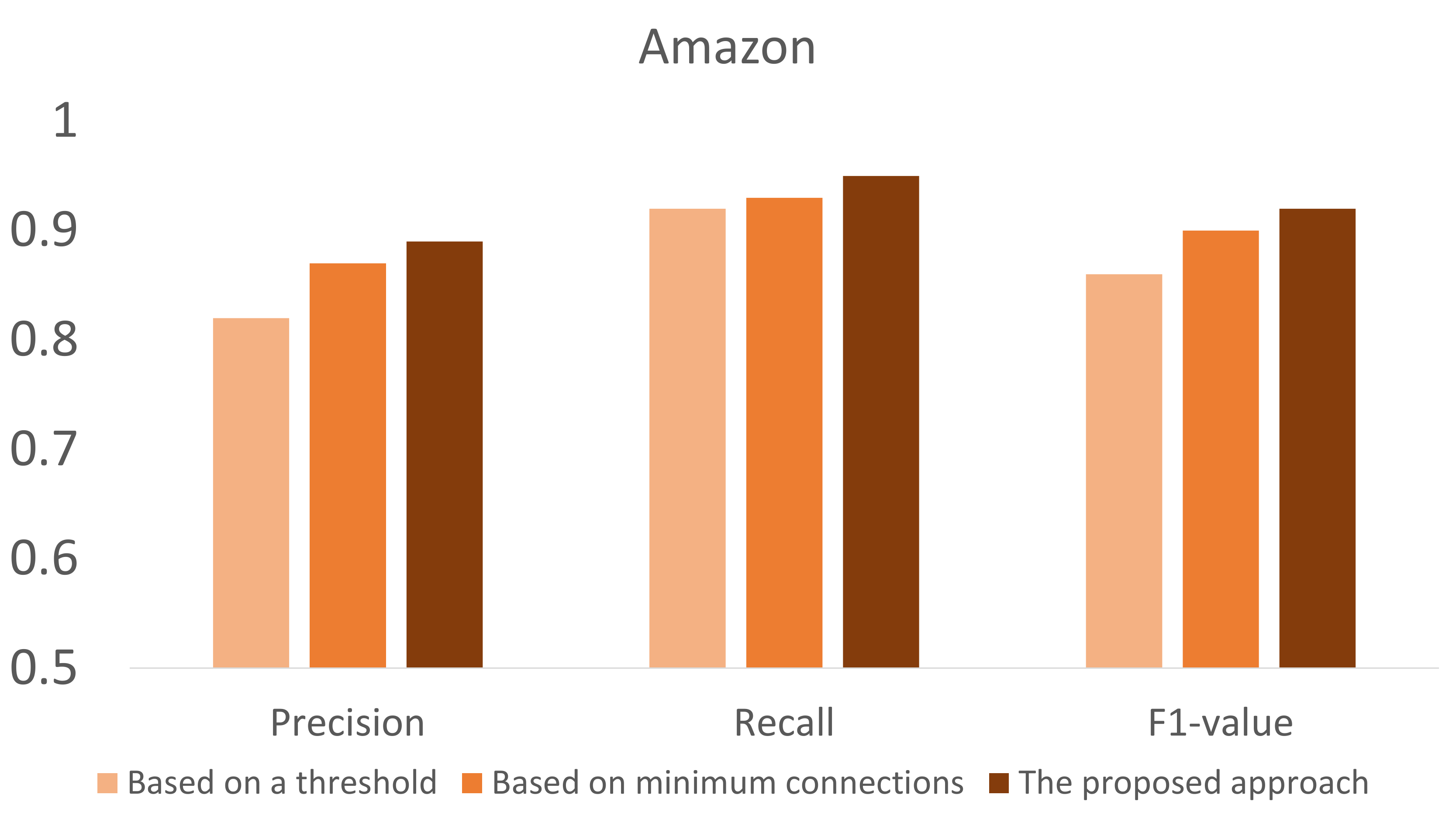}
\caption{The performance of the proposed approach for different removal strategy on the Amazon dataset.} 
\label{fig:clus-str-amazon}
\end{figure}
\textbf{Ablative Study:} 
The clustering removes fraudster reviewer(s) from a genuine group, and genuine reviewers from a fraudster group, 
thus reducing the $FP$. This in turn results in a better precision. On the other hand, with the removal of the genuine reviewer(s) from a fraudster group, the fraudster group's probability to be a fraudster is accordingly increased, thus reducing the $FN$. As a result of a decrease in $FN$, recall is also improved. However, the improvement in the F1-value for the Yelp dataset is significantly higher than the F1-value improvement on the Amazon dataset. This is mainly because the camouflage activity (the number of interactions between fraudster reviewers and genuine groups) in the Yelp dataset is higher.\\
\textbf{Comparative Analysis of Clustering:} 
To demonstrate the effectiveness of the proposed removal strategy, we also used two different strategies from previous studies:\\ 
\textbf{1)} Removing reviewers if their distances (Euclidean) from the centeroid are above a certain threshold~\cite{JI2020454} in the 1-means algorithm (only one centeroid).\\ 
\textbf{2)} Removing reviewers with minimum connections as suggested by Shehnepoor \textit{et al.}~\cite{shehnepoor2021hinrnn}.\\
Figs.~\ref{fig:clus-str-yelp} and \ref{fig:clus-str-amazon} display the performance of the proposed approach on different removal strategies. As shown the performance is improved against two other strategies. Using a threshold to remove the reviewers based on a representation distance from the centroid, likely removes some reviewers from a group, but not all groups are necessarily a mix of fraudsters and genuine reviewers.

Removing reviewers based on a minimum number of connections results in inaccurate removals. This is because some fraudsters make connections with multiple genuine reviewers in a group to escape detection. Similarly, genuine reviewers are also unintentionally involved in fraudster activities, because they happen to have reviewed the same item with the same rating and similar semantics. 
The proposed approach, on the other hand, utilizes the encoded spatio-temporal relations between reviewers and then refines the representations based on the given relations through the GCN. The proposed strategy for the clustering step will remove the outlier reviewer(s) in the feature space. If no outlier reviewers are recognized in a group then all the reviewers remain in the group. \\
\begin{figure}[ht]
\centering
\includegraphics[width=0.47\textwidth]{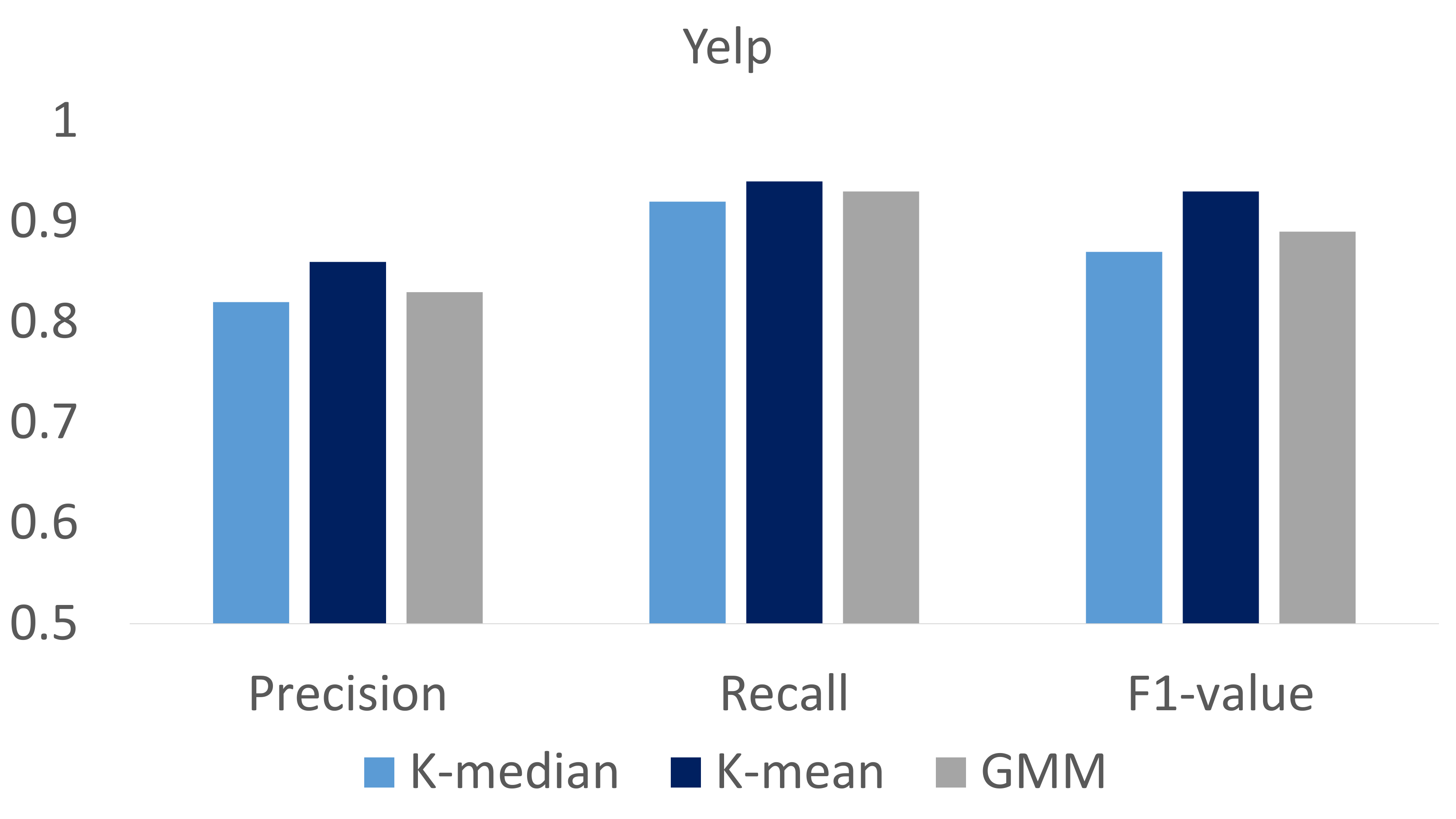}
\caption{The performance of K-mean and K-median on the Yelp dataset.} 
\label{fig:KNN-versions-yelp}
\end{figure}
\begin{figure}[ht]
\centering
\includegraphics[width=0.47\textwidth]{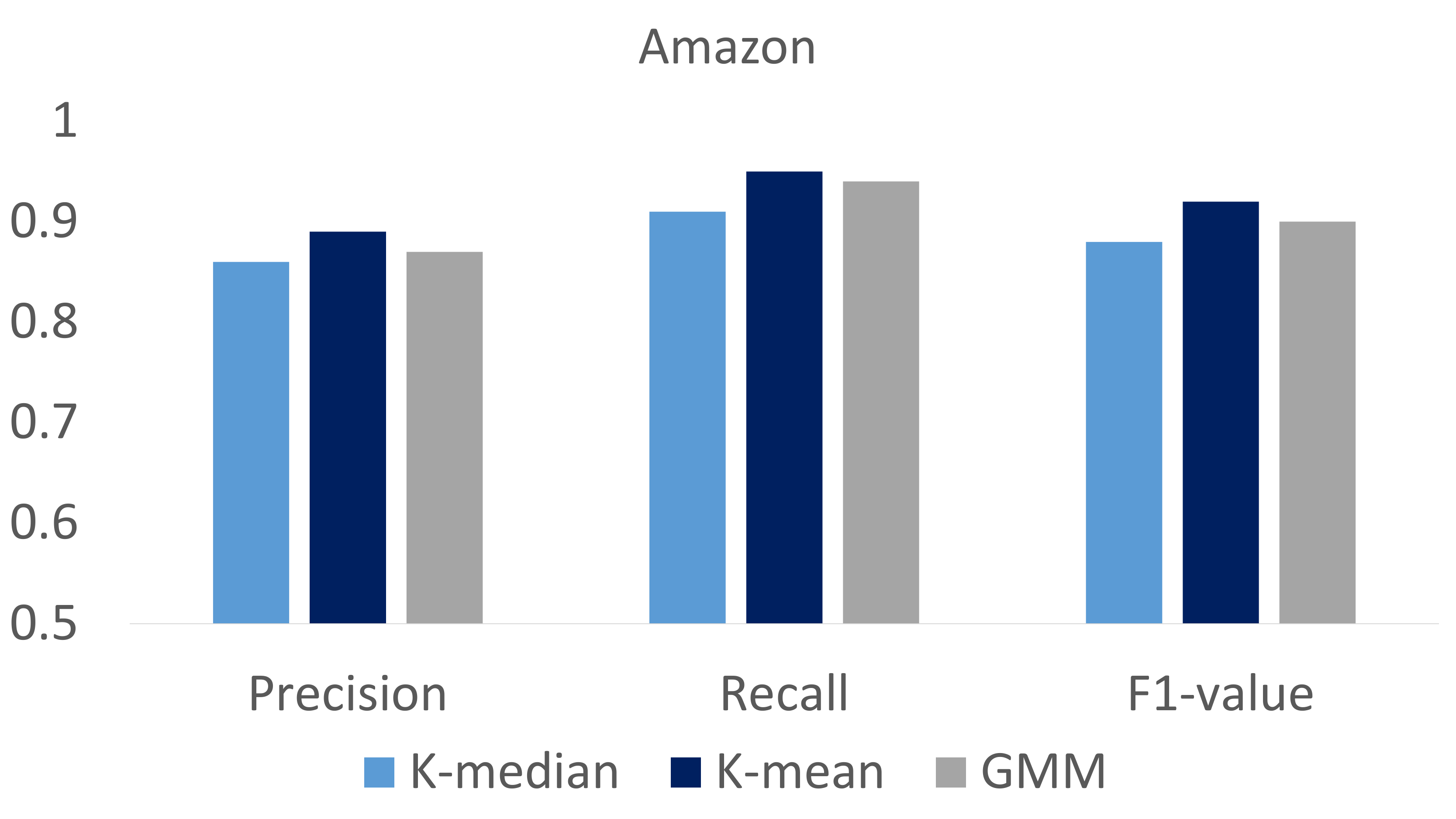}
\caption{The performance of K-mean and K-median on the Amazon dataset.} 
\label{fig:KNN-versions-amazon}
\end{figure}
\textbf{Observations:}
To compare the effectiveness of different clustering versions, we used three different clustering algorithm: \textbf{first,} we used Gaussian Mixture Model (GMM) as the clustering algorithm. In this algorithm, the reviewers with a distance greater than one standard deviation (Mahalanobis distance $>$ 1) were removed. As for the second algorithm we used the K-medians \cite{park2009simple}, with the median as the centroid. This, intuitively better represents the centroid, as the average of representations may not be a suitable centroid of a group, due to the existence of outlier reviewers. The second version utilizes the K-means \cite{hamerly2003learning}, as the proposed approach. 
Figs.~\ref{fig:KNN-versions-yelp} and \ref{fig:KNN-versions-amazon} show the performance of the proposed approach based on K-means, K-medians, and GMM. Using the mean to calculate the centroid of groups improves the performance of the proposed approach compared to the version which employed the median to obtain the centroid. Similarly using only the centroid from K-means demonstrated a better performance than using the GMM's Mahalanobis distance. As shown the improvement is the same almost for both datasets. 

\section{Conclusion}
\label{sec:conclusion}
Previous studies incorporated static networks to model a collaboration matrix on the fraudster group detection task, overlooking temporal features, while suffering from limitations in incorporating reviewers' influence on each others. In this study, we proposed a four-step framework to address these limitations: spatial modeling of the reviewers' co-review behaviors in different time windows, temporal modeling of spatial relations in a sequence of time windows, the reviewers' representation refinement through GCN, and the outlier removal from the groups through K-means clustering before the final classification of groups. The proposed approach effectively improved the performance of the proposed approach by 5\% (4\%), 12\% (5\%), 12 (5\%)\% on the Yelp (Amazon) dataset, respectively. For future work, the stochastic modeling of the reviewers' relations can be considered to provide a better behavioral representation of the reviewers, as most fraudsters randomly change their grouping relations to escape detection. 

\bibliographystyle{IEEEtran}
\bibliography{references.bib}

\end{document}